%% file: main.tex
\input{chapters/shared_preamble.tex}
\projectlinks{Project: \url{https://phy-agent-os.net} \quad GitHub: \url{https://github.com/PhyAgentOS/PhyAgentOS}}
\begin{document}
\sloppy
\maketitle

\input{chapters/abstract.tex}
\input{chapters/01_introduction/chapter.tex}

\input{chapters/02_related_work/chapter.tex}
\input{chapters/03_system_design/chapter.tex}

\input{chapters/04_core_mechanisms/chapter.tex}
\input{chapters/05_experiments/chapter.tex}

\input{chapters/06_conclusion/chapter.tex}
\input{chapters/07_appendix/chapter.tex}

\bibliographystyle{plainnat}
\bibliography{main}

\end{document}

%% file: chapters/shared_preamble.tex
\documentclass[11pt,a4paper]{deepseek}

\usepackage{amssymb}
\usepackage{longtable}
\usepackage{listings}
\usepackage[numbers,sort&compress]{natbib}
\usepackage{multirow}
\graphicspath{{figures/}{../../figures/}}

\title{PhyAgentOS: A Self-Evolving Operating System for Embodied Agents with Decoupled Cognitive Planning and Physical Execution}

\author{
    Yang Liu\textsuperscript{1,2}, Weixing Chen\textsuperscript{2}, Xinshuai Song\textsuperscript{2}, Tao Pu\textsuperscript{1}, Siwen Mo\textsuperscript{2}, Yongjie Bai\textsuperscript{2,3}, Zihao Chen\textsuperscript{2,3}, Qianran Sun\textsuperscript{2}, Liruo Zhong\textsuperscript{1}, Ying Shen\textsuperscript{1,2}, Liang Lin\textsuperscript{1,2,3,\dag}
}

\correspondingauthor{\dag\ Corresponding author}

\institutionlogos{%
    \begin{minipage}[t]{0.28\textwidth}
        \centering
        \includegraphics[height=1.0cm]{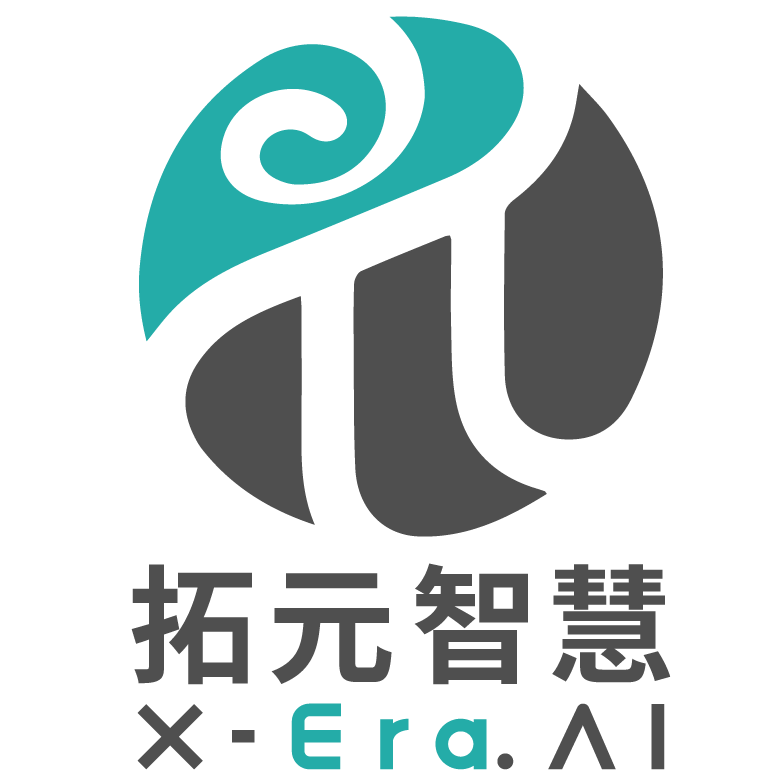}\\[2pt]
        {\small\color{mornavy} \textsuperscript{1}X-Era Lab}
    \end{minipage}%
    \hfill
    \begin{minipage}[t]{0.36\textwidth}
        \centering
        \includegraphics[height=1.0cm]{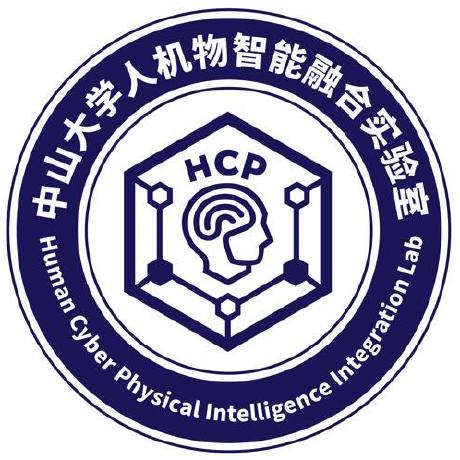}\\[2pt]
        {\small\color{mornavy} \textsuperscript{2}HCP Lab, Sun Yat-sen University}
    \end{minipage}%
    \hfill
    \begin{minipage}[t]{0.28\textwidth}
        \centering
        \includegraphics[height=1.0cm]{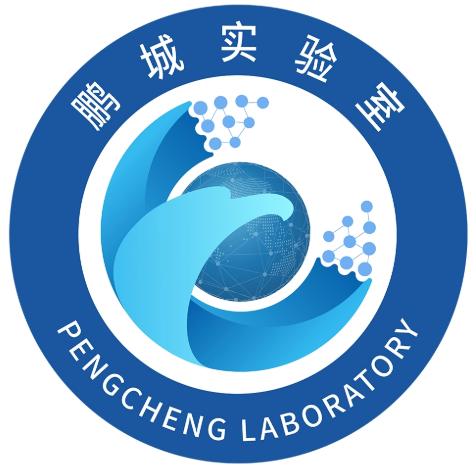}\\[2pt]
        {\small\color{mornavy} \textsuperscript{3}Peng Cheng Laboratory}
    \end{minipage}%
}

\version{v0.1.6}

\keywords{Embodied AI, Agentic workflow, Cognitive-physical decoupling, Self-evolution, Robot operating system, Language-action interface}

%% file: chapters/abstract.tex
\begin{abstract}
Vision-language-action models, world models, and agentic planners each address important aspects of physical intelligence. However, their composition alone does not provide a common execution abstraction, shared state representation, semantic verification, or persistent experience across heterogeneous embodiments. 
ROS provides message-passing infrastructure, whereas task-level orchestration and outcome interpretation remain application specific. We present \textbf{PhyAgentOS}, a runtime foundation that delivers scheduling, verification, memory, benchmarking, and safety as system-level services for physical agents.
PhyAgentOS is organized around a \textbf{Session-Centered Runtime}, in which a session rather than an individual action is treated as the minimum unit of scheduling, compatibility preflight, supervised execution, evidence collection, and acceptance.
To decouple cognitive planning from physical execution, the system represents the cognition-physics boundary as a file system. 
Its \textbf{State-as-a-File protocol} materializes cross-layer state in human-readable Markdown documents with embedded YAML, producing an inspectable and versionable execution record without requiring direct code dependencies between the Agent and Runtime layers. 
These protocol views jointly form a unified cognitive state space that aligns task intent, runtime and target capabilities, structured environment state, execution status, and historical experience.
This shared state enables the \textbf{SessionVerifier} to distinguish execution termination from semantic task completion by issuing evidence-grounded verdicts of success, failure, or replan.
Verified outcomes are consolidated through epistemic memory into reusable knowledge and corrective lessons, forming a closed trial-and-error loop across sessions without requiring neural-model retraining. 
Benchmarking reuses the same session, runtime, and verification path as deployment, making evaluation results traceable to the actual execution process. 
Both policy-driven and Agent-driven execution are constrained by a layered safety architecture that is composed of compatibility preflight, action bridges, \texttt{SafetyGuard}, heartbeat monitoring, and target-local constraints.
Validation follows a progressive path in which game environments are used to evaluate cognitive planning without low-level control, simulation introduces physical dynamics, latency, and embodied control challenges, and real robots bring in hardware constraints, environmental noise, and actuator errors, while the cognitive layer is held constant across all tiers.
PhyAgentOS is benchmarked on Optimus-67, StarDojo, and DST-Dojo for cognitive planning, and its unified protocol is validated on more than 19 robot embodiments across simulated and physical platforms. Further evaluation on LIBERO, Calvin and RoboCasa365 demonstrates system usability and performance gains across multiple vision-language-action models.
\end{abstract}

%% file: chapters/01_introduction/chapter.tex
\section{Introduction}
\label{sec:intro}

When a robot is instructed to grasp a cup, its gripper may close on empty air. The VLA controller records that the trajectory was completed within tolerance, the world model confirms agreement between its predicted state and the observation, and the agentic planner reports that all tool calls returned without error. Every layer signals success, yet the cup is not in the robot’s hand. This is not merely a corner case, it reflects a common structural failure in contemporary embodied AI systems. No architectural layer mediates between high-level intent and physical action to verify whether the intended outcome has been realized, diagnose why it was not, or prevent recurrence.

This verification gap constitutes a fundamental limitation of current architectures. The pipeline from natural-language instruction to motor command is largely designed as a unidirectional process: the planner decomposes the goal, the model generates actions, and the actuator tracks the resulting trajectory. However, no component is explicitly responsible for semantically verifying the physical outcome. As a result, execution termination is often conflated with task completion, and a successful controller return code is interpreted as evidence of goal achievement~\cite{pettersson2005execution,ingrand2017deliberation, sliwowski2025conditionnet}. This conflation produces systematic self-deception: an execution trace may terminate normally even as the resulting environmental state fails to satisfy the intended objective.

\begin{figure}[htbp]
    \centering
    \includegraphics[width=0.95\textwidth]{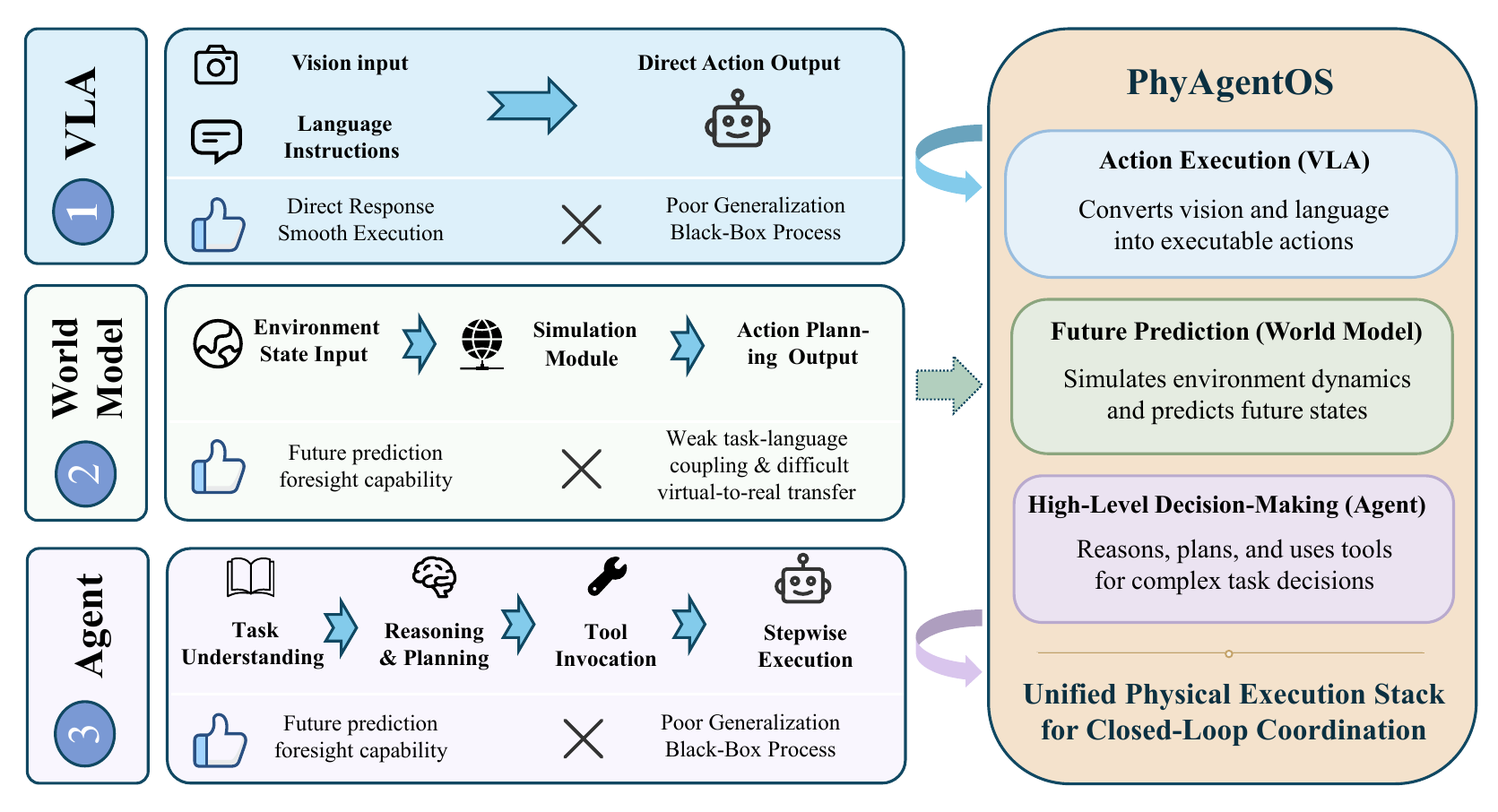}
    \caption{Three competing visions for AI in the physical world: VLA models (direct perception-to-action), world models (predictive simulation), and agentic systems (high-level reasoning and tool use). Each excels in isolation, but none provides the architectural layer that verifies execution, persists memory, and isolates faults across paradigms.}
    \label{fig:three_visions}
\end{figure}

The verification gap also arises from the field’s organization around largely distinct paradigms, each addressing a particular subproblem in relative isolation~\cite{liu2025aligning}. Vision-language-action models map perceptual inputs directly to motor commands and have demonstrated impressive end-to-end generalization~\cite{jang2022bcz,brohan2023rt1,brohan2023rt2,openx2024rtx,bousmalis2023robocat,ghosh2024octo,kim2024openvla,black2025pi0,bai2026tvve, cheng2025navila,bu2025univla,hou2025dita}. However, they typically provide limited diagnostic feedback regarding task outcomes. World models learn predictive dynamics from interaction data to support look-ahead planning~\cite{hafner2019planet,hafner2020dreamer,hafner2025dreamerv3,wu2023daydreamer,hansen2024tdmpc2,bruce2024genie,zhou2024robodreamer}, but they are often only weakly coupled to high-level task language. Agentic systems decompose tasks, reason about tool availability, and perform multi-step inference~\cite{ahn2022saycan,huang2022inner,driess2023palm,liang2023codeaspolicies,singh2023progprompt,song2023llmplanner,huang2023voxposer,wen2024rekep}; nevertheless, planning and execution typically remain within the same process boundary. Although each paradigm has made meaningful progress in its respective domain, none is inherently responsible for verifying physical task outcomes. Consequently, when these components are integrated into a single pipeline, the verification function remains unassigned. Figure \ref{fig:three_visions} illustrates this architectural separation: three distinct pipelines—perception-to-action, predictive simulation, and high-level reasoning—operate without a shared state representation or common verification layer. The unified PhyAgentOS framework, shown on the right, is introduced in the following sections as an integrative layer designed to address this gap.

The straightforward composition of an agentic planner, a VLA controller, and a world model for look-ahead rollouts fails for three structural reasons. First, the layers operate at incompatible levels of abstraction. A VLA model processes pixel arrays and text tokens and outputs joint-level trajectories, whereas a world model operates on compressed latent-state representations and an agentic planner reasons over symbolic tool calls. Because no shared state representation bridges these heterogeneous formats, ad hoc converters are required at each interface. Second, opacity is compounded rather than mitigated. When a ten-step plan interleaves multiple VLA calls with world-model rollouts and fails at an intermediate step, identifying the root cause requires searching an expanded space spanning reasoning, tool invocation, and physical execution. Third, the stack lacks persistent memory. Knowledge acquired during one session, such as a grasping strategy that succeeds for a particular object geometry or a navigation route that avoids a known obstacle, is discarded when the session ends. Consequently, the combined stack amplifies the limitations of its individual components rather than compensating for them.

\begin{figure}[htbp]
    \centering
    \includegraphics[width=0.95\textwidth]{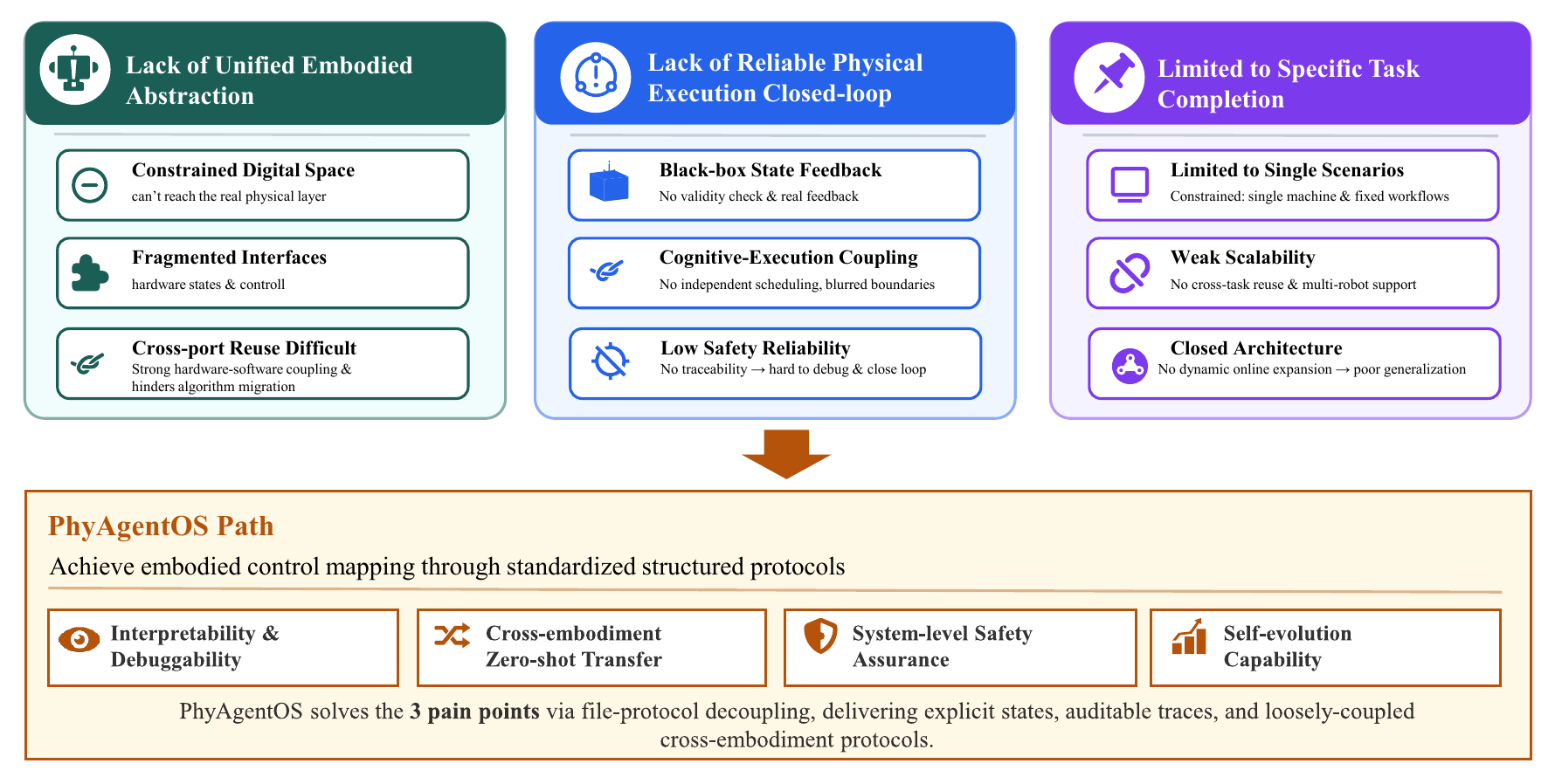}
    \caption{Three requirements for an embodied AI operating system: unified embodiment interface (left), reliable execution and verification loop (center), and cross-task learning (right). PhyAgentOS addresses each requirement through explicit file protocols and adapter chains, a Session-Centered Runtime with supervisory scheduling, and self-evolution mechanisms driven by multi-tier persistent memory.}
    \label{fig:three_requirements}
\end{figure}

A common misconception is that a robot communication layer is equivalent to an operating system. Supervisory robot control, layered autonomy, behavior trees, and integrated task-and-motion planning offer complementary abstractions for execution~\cite{simmons1994structured, alami1998architecture, iovino2022behaviortrees, garrett2021tamp}. Although the Robot Operating System (ROS) is the most widely deployed software infrastructure in robotics, it does not provide the missing supervisory layer. ROS primarily supplies publish-subscribe communication mechanisms~\cite{quigley2009ros,macenski2022ros2}, serving as a communication backbone rather than an embodied-AI operating system. It does not natively provide task-level scheduling, persistent cross-session memory, semantic verification of task completion, or a unified layered safety architecture. Instead, safety mechanisms are typically implemented separately, often relying on low-level safeguards such as emergency stops. ROS therefore functions as middleware that connects sensors, actuators, and software components. An operating system for embodied AI must operate above this middleware and provide scheduling, memory, verification, and safety as first-class services. Figure \ref{fig:three_requirements} formalizes the gap between the capabilities provided by existing middleware and those required for embodied AI. The upper row identifies three structural deficiencies shared across the three paradigms, whereas the lower row presents the corresponding capabilities that an operating-system layer must provide.

We present PhyAgentOS, an operating system that sits beneath these paradigms. Its central design choice is to treat the cognition-physics boundary as a file system rather than a function-call interface. Every cross-process datum, including targets, skills, sessions, observations, lessons, and knowledge, is materialized in human-readable Markdown files with embedded YAML blocks. This State-as-a-File protocol yields an append-only audit trail by construction, 
and supports language-agnostic interaction.

This protocol supports the Session-Centered Runtime. Sessions, not atomic actions, are the unit of scheduling, preflight validation, runtime monitoring, and post-hoc verification. The WatchdogSupervisor claims pending sessions from SESSIONS.md, validates adapter contracts, monitors execution heartbeats, and writes results back. Cognition and physics are separated into independent processes; neither imports the implementation of the other. The protocol files collectively constitute a multi-tier memory architecture, the persistent substrate on which the system learns. SESSIONS.md stores episodic memory. ENVIRONMENT.md provides working memory. LESSONS.md and KNOWLEDGE.md form semantic memory. SKILL.md and SKILLRUNTIME.md encode procedural memory. Continual robot learning studies retention and transfer across tasks~\cite{thrun1995lifelong,lesort2020continualrobotics,liu2023libero}, while skill libraries and verbal reflection enable language agents to reuse experience without updating model weights~\cite{wang2023voyager,shinn2023reflexion,zhao2024expel}. This multi-tier memory drives a self-evolution loop across sessions, tasks, and embodiments without retraining any neural model.

Prior work on feedback-driven replanning, uncertainty-aware intervention, failure monitoring, and learned visual rewards motivates semantic outcome assessment~\cite{huang2022inner, ren2023knowno, guo2024doremi, liu2023reflect, sermanet2024robovqa, zhou2025codeasmonitor, ma2023liv, rocamonde2024vlmrewards, sontakke2023roboclip}. 
These efforts are systematized through five mechanisms that convert architectural guarantees into reliable physical behavior. 
First, the protocol layer is used to establish a unified cognitive state space. Cross-layer state is materialized in human-readable Markdown documents with embedded YAML, and task intent, runtime capabilities, environment state, execution status, and historical experience are thereby aligned in a single reference frame. Second, the SessionVerifier is built on this shared state to replace the binary "action finished" signal with semantic acceptance. Verdicts of success, failure, or replan are rendered from an evidence bundle rather than from controller return codes. Third, verified outcomes are consolidated through epistemic memory into reusable knowledge and corrective lessons, which closes the trial-and-error loop across sessions without requiring neural-model retraining. Fourth, benchmarking is treated as an instrument for self-evolution rather than as an external script. The same session, runtime, and verification path used in deployment is reused during evaluation, so measured gains are traceable to the actual execution process. Fifth, both policy-driven and agent-driven execution are constrained by a layered safety architecture that is composed of compatibility preflight, action bridges, SafetyGuard, heartbeat monitoring, and target-local constraints.


Validation follows a progressive path. Game environments strip away physics and execution noise, isolating memory, planning, and self-evolution as the objects of study. Simulation adds rigid-body dynamics, collision detection, and configurable latency. Real robots add hardware noise, sensor uncertainty, and safety-critical constraints. The cognitive layer remains unchanged across all three tiers. Across Optimus-67, StarDojo, and DST-Dojo benchmarks, PhyAgentOS demonstrates that a single protocol infrastructure and supervisory runtime can drive diverse embodiments from game engines to humanoid robots. At the endpoint of this pipeline, a successful grasp carries an evidence bundle that proves it; a failed grasp is recorded, attributed, and prevented from recurring.

The remainder of this paper is organized as follows. Section 2 reviews related work. Section 3 presents the three-layer architecture. Section 4 details the five core mechanisms. Section 5 describes the progressive validation pipeline, benchmark results, and ablation studies. Section 6 concludes with limitations and future work.

%% file: chapters/02_related_work/chapter.tex
\section{Related Work}
\label{sec:related}

PhyAgentOS lies at the intersection of three active research streams (vision-language-action models, world models, and agentic systems) and draws on a fourth (embodied AI frameworks and robot Communication Infrastructure). We review each stream, identify its characteristic limitations, and position our system relative to the dominant approaches.

\subsection{Vision-Language-Action Models}

Vision-language-action (VLA) models learn end-to-end policies that map visual observations and natural language instructions directly to robot actions~\cite{ma2026vlasurvey}. Task-conditioned manipulation and generalist policies have progressed from CLIPort, BC-Z, Gato, and RT-1 to cross-robot RT-X, RoboCat, and Octo~\cite{shridhar2021cliport,jang2022bcz,reed2022gato,brohan2023rt1,openx2024rtx,bousmalis2023robocat,ghosh2024octo}. RT-2~\cite{brohan2023rt2} and OpenVLA~\cite{kim2024openvla} demonstrate that large-scale vision-language pretraining enables strong generalization and effective fine-tuning for manipulation tasks, while recent work further improves action generation and scalability through alternative modeling paradigms~\cite{zhao2023act,chi2025diffusionpolicy,liu2025rdt,belkhale2024rth,pertsch2025fast, zhao2025cotvla,shi2025hirobot,wang2025vqvla,li2025coavla,zhou2025chatvla,li2026pointvla,wang2026vlaadapter,shukor2025smolvla,romer2026clare,zhan2026stablelanguage,li2026tcot}. In particular, flow- and diffusion-based methods such as ${\pi}_0$~\cite{black2025pi0} and the recent ${\pi}_{0.5}$~\cite{intelligence2025pi05} model action trajectories as generative processes to enhance multimodality and temporal consistency. GR00T N1 explores a generalist humanoid foundation model trained on heterogeneous robot data~\cite{nvidia2025gr00t}. Recent systems improve spatial action representations, couple vision-language backbones to diffusion experts, and scale unified diffusion-autoregressive policies~\cite{qu2025spatialvla,wen2025dexvla,wen2025diffusionvla}. Gemini Robotics reports open-vocabulary manipulation, robustness to object and environment variation, and adaptation to new embodiments~\cite{geminirobotics2025}; cross-embodiment scaling and reinforcement fine-tuning with verified simulator rewards remain active directions~\cite{zheng2025xvla,li2025vlarft}.


These models share a structural opacity. When an action fails, the monolithic architecture offers little diagnostic signal about whether the error stems from visual misperception, policy misgeneralization, or execution dynamics. The model produces a motor command; the operator receives a success or failure outcome, with negligible intermediate attribution. Recent robustness-oriented robotic manipulation research addresses distribution shifts through complementary strategies, including consistency regularization across instructions, trajectories, and observations, as well as task-aware viewpoint selection under occlusion and camera-pose variations~\cite{luo2026rovla,bai2026tvve}. Deployment, however, remains sensitive to data coverage and embodiment or viewpoint shifts~\cite{khazatsky2024droid,li2024simpler,kim2025openvlaoft}, as well as the compounding-error and causal-confusion failure modes of behavior cloning~\cite{ross2011dagger,dehaan2019causal}. These brittlenesses motivate a system architecture in which the VLA is treated as a replaceable component rather than the sole decision-making pathway.
PhyAgentOS integrates VLAs as swappable \texttt{PolicySkillRuntime} backends rather than as monolithic controllers. The framework handles observation normalization, action chunking, and safety clamping around the policy; the VLA remains an interchangeable module whose outputs are validated and patched before they reach the physical target. This design isolates the policy from the execution infrastructure, enabling a single VLA to be deployed across multiple embodiments through adapter chains, and allowing failures to be attributed to the policy, the adapter, or the target separately.

\subsection{World Models}

World models learn compressed representations of environment dynamics to predict future states conditioned on actions, enabling model-based look-ahead planning~\cite{hafner2019planet,hafner2020dreamer,hafner2021dreamerv2,hafner2025dreamerv3,schrittwieser2020muzero,hansen2024tdmpc2,moerland2023mbrlsurvey,finn2017visualforesight,dasari2019robonet,wu2023daydreamer, guo2026flowdreamer,zhu2025irasim}. Modern approaches increasingly explore generative world simulation. Genie learns action-controllable generative environments~\cite{bruce2024genie}, UniPi uses text-guided video generation for policy learning~\cite{du2023unipi}, and RoboDreamer learns compositional world models for robot imagination~\cite{zhou2024robodreamer}; Envisioner extends generative world modeling to robotic manipulation~\cite{liao2025genieenvisioner}. Recent robot world models also support task-agnostic test-time planning in pretrained feature spaces~\cite{zhou2025dinowm} and synthesize neural trajectories for behavior and environment generalization~\cite{jang2025dreamgen}. Meanwhile, universal simulation frameworks such as UniSim~\cite{yang2023unisim} aim to model diverse embodied interaction scenarios across robotic embodiments. Beyond simulation, recent work leverages learned or generated dynamics for data synthesis and policy optimization. RoboGen~\cite{wang2023robogen} repurposes generative simulation for synthetic data generation in robotic learning, while related approaches explore using learned dynamics models for reinforcement learning fine-tuning and scalable data augmentation. Generative simulation can also create tasks and reward functions through systems such as GenSim and Eureka~\cite{wang2024gensim,ma2024eureka}. Recent world-action modeling research further bridges prediction and control by integrating vision-language action proposals, action-conditioned future prediction, rollout-consistency verification, and compact future-change representations into decision making~\cite{zhao2024vlmpc,liu2026wav,bai2026bridgewa}. These approaches improve future-aware control and predictive reliability, but remain model-level mechanisms rather than system-wide solutions for execution contracts, post-execution semantic verification, or persistent cross-session memory.


Despite these advancements, deploying neural world models as standalone control substrates faces several system-level challenges. First, they are weakly coupled to abstract task instructions, often requiring intermediate semantic representations or external task decomposition mechanisms to bridge the gap between high-level language goals and continuous control signals~\cite{du2023unipi}. Second, sim-to-real transfer remains brittle due to compounding temporal prediction errors~\cite{janner2019mbpo,moerland2023mbrlsurvey} and unmodeled hardware noise, which limits robust generalization across heterogeneous robotic embodiments without extensive realignment~\cite{tobin2017domainrandomization,peng2018dynamicsrandomization,chebotar2019simopt}.

PhyAgentOS does not compete with world models; instead, it provides the runtime infrastructure required to make their predictions actionable. Through modular target adapters, the framework standardizes heterogeneous state spaces into unified observation formats consumable by generative world models. Conversely, policy adapters convert the model's predicted video trajectories or symbolic rollouts into structured session protocols. This architecture treats world models as plug-and-play forecasting engines within a closed-loop system, scheduled by the \texttt{WatchdogSupervisor} and validated by the \texttt{SessionVerifier} before any predictive commands are executed on physical hardware.

\subsection{Agentic Systems}

Agentic systems employ large foundation models to decompose high-level instructions into structured plans, reason over tool availability, and chain multi-step inference~\cite{yao2023react,schick2023toolformer,huang2022zeroshotplanner,liang2023codeaspolicies,singh2023progprompt,song2023llmplanner, yang2025magma,ji2025robobrain,wang2025karma}. Early implementations like SayCan~\cite{ahn2022saycan} grounded LLM-generated plans via robot affordances, scoring feasibility alongside semantic coherence. Inner Monologue~\cite{huang2022inner} and PaLM-E~\cite{driess2023palm} advanced this paradigm by injecting real-time environmental feedback and multi-modal sensor streams directly into the core backbone for closed-loop replanning. More recently, frameworks like Voyager~\cite{wang2023voyager} and RoboAgent~\cite{bharadhwaj2023roboagent} introduce lifelong skill libraries, continuous manipulation optimization, and policy-level reflection mechanisms, treating low-level controllers as evolving tools to solve long-horizon, complex sequential tasks. Uncertainty-aware help-seeking and experiential reflection further support intervention, correction, and reusable knowledge~\cite{ren2023knowno,shinn2023reflexion,zhao2024expel,wang2023deps}. AutoRT demonstrates foundation-model-assisted orchestration of a robot fleet for scalable in-the-wild data collection~\cite{ahn2024autort}. Embodied Agent Interface formalizes common LLM decision modules and fine-grained error metrics, while PARTNR exposes persistent task-tracking, coordination, and error-recovery failures in collaborative embodied agents~\cite{li2024embodiedinterface,chang2025partnr}.

A shared weakness unites these systems. Planning and execution typically reside within a monolithic process or single API boundary. When an execution fails, it remains ambiguous whether the high-level LLM made a poor decision, an intermediate tool call was mishandled, or a physical controller failed to track the command. This tight coupling forces hardware interfaces to be re-implemented for each new robot and complicates debugging across the cognition–execution divide, while compounding inference latencies make long-horizon tracking brittle.

PhyAgentOS adopts a \emph{control tower} model for the Agent layer. The Agent observes incoming requests, fuses context from the protocol files (environment state, session history, lessons, knowledge), calls tools through a constrained \texttt{ToolRegistry}, and emits structured sessions into \texttt{SESSIONS.md}. Actual execution is delegated entirely to the Runtime layer. This decoupling means that the Agent reasons about \emph{what} to do, while the \texttt{WatchdogSupervisor}, \texttt{SessionRunner}, and \texttt{SkillRuntime} handle \emph{how} it gets done. Failures can be attributed to the planning phase, the execution phase, or the adapter chain independently, and the same Agent logic can drive multiple embodiments simply by targeting different \texttt{BaseRolloutTarget} instances through the same file protocol.

\subsection{Embodied AI Frameworks and Robot Communication Infrastructure}

A growing ecosystem of simulation platforms and benchmarks has driven progress in embodied AI by providing reproducible physics and standardized evaluation protocols. While environments like Habitat~\cite{szot2021habitat} and Gibson~\cite{li2021igibson} target high-fidelity indoor navigation and interaction, RLBench~\cite{james2019rlbench}, RoboSuite~\cite{zhu2025robosuite}, and ManiSkill~\cite{mu2021maniskill} explicitly focus on continuous, large-scale articulated manipulation. Complementary frameworks such as Isaac Gym~\cite{makoviychuk2021isaacgym} and OpenAI Gym~\cite{brockman2016openaigym} provide standardized simulation and reinforcement learning interfaces, enabling scalable benchmarking and reproducible training across diverse robotic platforms.

On the hardware side, the Robot Operating System (ROS) and manufacturer-specific SDKs provide mature low-level control primitives, message-passing infrastructure, and hardware abstraction. Yet an explicit, auditable boundary between high-level cognition and physical execution is not provided. Task specification, planning, and execution monitoring are typically left to application-level code built ad-hoc on top of the communication layer, and no standardized protocol is provided for decoupling the cognitive agent from the physical target.

PhyAgentOS unifies game engines, simulators, and real-robot hardware behind a single \texttt{BaseRolloutTarget} interface and a shared session protocol. Each target exposes a standardized set of capabilities (observe, execute, reset) through adapter chains, while the \texttt{WatchdogSupervisor} schedules sessions identically regardless of whether the target is a Stardew Valley game instance, a MuJoCo simulation, or a Franka arm. This abstraction enables a single experimental pipeline to validate cognitive strategies in low-cost game environments, transfer them to simulation for dynamics verification, and deploy them on real hardware, with the same protocol files serving as the audit trail across tiers.

\subsection{Embodied Agent Systems}

Recent studies have extended embodied intelligence from individual policy models toward complete agent systems that integrate reasoning, capability invocation, execution feedback, memory, and safety. Unlike model-centric approaches that focus primarily on observation-to-action prediction, embodied agent systems investigate how high-level goals are translated into persistent and controllable interaction with heterogeneous physical environments.

One line of work formulates embodied control as executable program generation. CaP-X introduces a unified framework for benchmarking and improving Code-as-Policy agents and shows that high-level robot primitives can simplify planning but may also constrain policy expressiveness \cite{fu2026capx}. ASPIRE further treats robot programming as an iterative process of skill discovery, execution feedback, and program refinement, enabling validated solutions to be accumulated in an expandable skill library \cite{lu2026aspire}. These methods improve adaptability through closed-loop debugging, but remain closely tied to predefined perception and control primitives.

A complementary line emphasizes the runtime infrastructure connecting foundation models with robots. ROSClaw provides dynamic ROS 2 capability discovery, multimodal observation normalization, action validation, and execution auditing \cite{cardenas2026rosclaw}. RoboOS adopts a hierarchical architecture that combines global planning, embodiment-specific skills, and shared memory for cross-embodiment and multi-robot collaboration \cite{tan2025roboos}. ABot-Claw further incorporates capability scheduling, persistent multimodal memory, and critic-guided replanning for cooperative and self-evolving robotic agents \cite{huo2026abotclaw}. Collectively, these systems demonstrate the importance of capability abstraction, runtime supervision, and long-term feedback. However, a unified cognitive state space that aligns task intent, environment state, execution evidence, and historical experience across layers is not established. Semantic verification of physical outcomes against task objectives is not systematically performed. Trial-and-error cycles are not closed by verified consolidation into reusable knowledge. Benchmarking is typically separated from the deployment runtime, and safety is handled by isolated mechanisms rather than as a layered architecture. PhyAgentOS instead organizes these capabilities around a shared session and protocol abstraction. The same unified cognitive state space, evidence-grounded SessionVerifier, epistemic memory loop, deployment-identical benchmarking path, and defense-in-depth safety stack are thereby enabled to operate across games, simulators, and real robots.

%% file: chapters/03_system_design/chapter.tex
\section{System Design}
\label{sec:design}

PhyAgentOS is designed around a system-level question that is not addressed by a policy model alone: how can high-level intent be translated into physical interaction that is executable, interruptible, verifiable, and reusable across heterogeneous targets? A vision--language--action model may generate an action sequence, and an LLM-based agent may decompose a long-horizon goal, but neither capability by itself defines who is allowed to execute an action, how compatibility is checked, how failures are contained, or how execution evidence is returned for subsequent reasoning. PhyAgentOS therefore treats embodied intelligence as a coordinated runtime process rather than as a single model invocation.

The architecture separates this process into two reasoning planes. The Agent layer determines \emph{what should be done}: it interprets requests, constructs task context, decomposes goals, selects an execution method and target, and evaluates the returned evidence. The Runtime layer determines \emph{how the task is carried out}: it validates the execution contract, supervises the session lifecycle, maintains the observation--action loop, and enforces target-side safety. A shared protocol boundary connects the two planes without introducing direct code dependencies. This ordering---intent formation, controlled execution, and explicit state exchange---provides the organizing logic for the remainder of this section.

\subsection{Design Rationale}

The first design objective is to decouple cognitive reasoning from physical control. The Agent may use large models, external tools, and long-term memory, whereas latency-sensitive control must remain close to the target and obey target-specific constraints. PhyAgentOS therefore prohibits the Agent from issuing raw hardware commands. Instead, the Agent produces a structured \emph{session} that specifies the task objective, selected \texttt{SkillRuntime}, target, preconditions, and acceptance criteria. The Runtime independently claims and executes that session. This separation allows either plane to evolve without requiring the other to import its implementation.

The second objective is to make the session, rather than an individual action, the minimum unit of system execution. Physical tasks unfold over time and cannot be governed safely by validating isolated commands alone. A session provides a bounded lifecycle within which compatibility checks, heartbeats, cancellation, retries, evidence collection, and result writeback can be applied consistently. Correspondingly, the \texttt{WatchdogSupervisor} performs supervision rather than control: it governs the lifecycle of a session, while the step-wise observation--action loop is delegated to downstream runtime components.

The third objective is to make cross-layer state explicit. All decisions that cross the Agent--Runtime boundary are represented by human-readable and machine-parseable protocol documents. This choice is not merely an implementation convenience. It turns execution into an inspectable state transition system: the requested goal, selected capability, target constraints, environment snapshot, terminal evidence, and recovered lesson can be examined independently, versioned, and replayed. Auditability and reproducibility therefore follow from the architecture rather than being added as logging features after the fact.

Finally, the system must preserve portability without erasing physical differences. Game environments, simulators, and real robots share a common session abstraction, but they expose different observations, action spaces, timing requirements, and safety constraints. PhyAgentOS isolates these differences behind target and policy adapters while retaining local safety enforcement at the target boundary. The same cognitive workflow can consequently be reused across embodiments, whereas the physical assumptions remain explicit and target-specific.

\subsection{Overall Architecture}

\begin{figure}[htbp]
    \centering
    \includegraphics[width=0.98\textwidth]{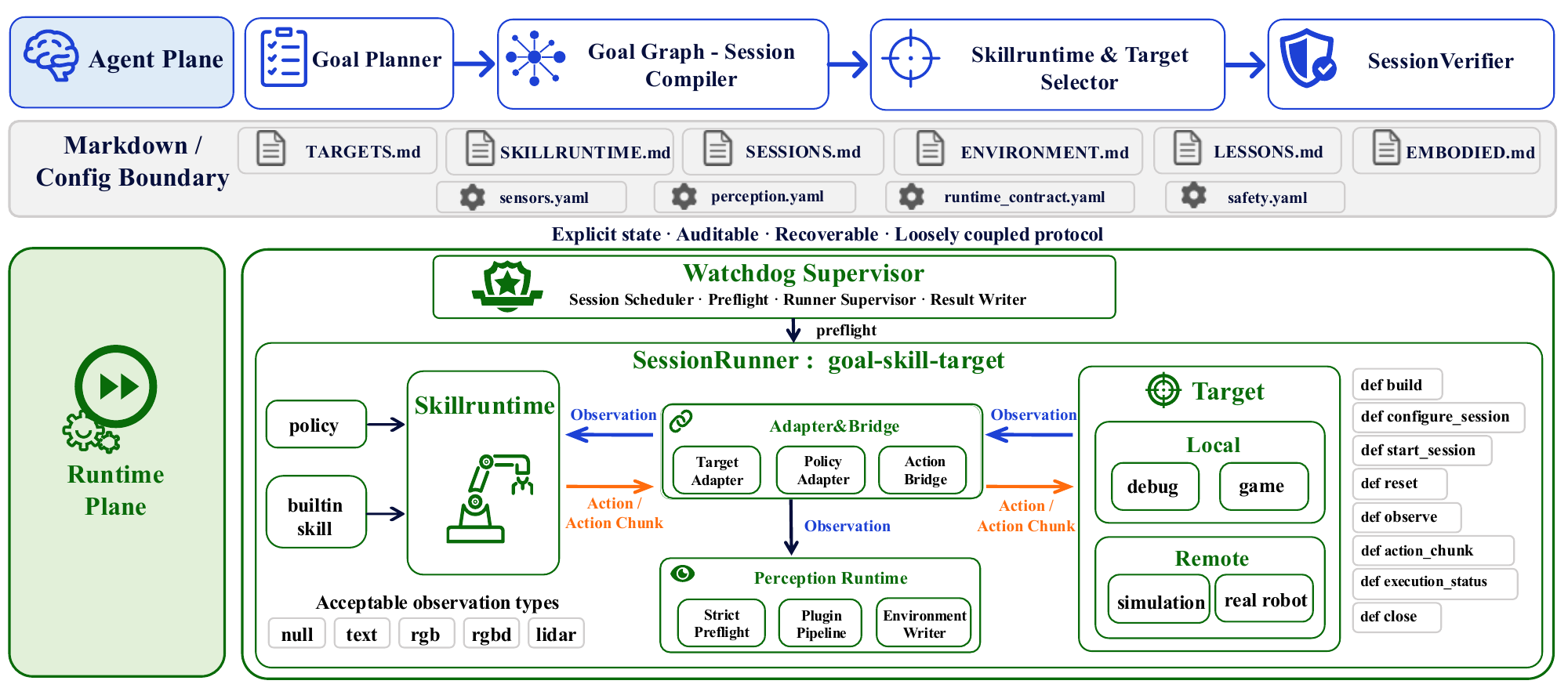}
    \caption{Overall architecture of PhyAgentOS. The agent layer determines task intent and compiles executable sessions, whereas the runtime layer supervises and performs target interaction. The shared protocol boundary makes the handoff explicit, auditable, and loosely coupled.}
    \label{fig:overall_architecture}
\end{figure}

Figure~\ref{fig:overall_architecture} presents PhyAgentOS as an agent layer and a runtime layer connected by a shared protocol boundary. The separation is functional rather than merely organizational. The upper plane converts an open-ended user request into a verifiable execution contract; the lower plane converts that contract into a bounded interaction with a game, simulator, or robot. The protocol boundary records the handoff and the returned evidence, thereby preventing either plane from depending on the internal control flow of the other.

An end-to-end task proceeds as follows. A request is first interpreted by the Goal Planner and decomposed by the Goal Graph and Session Compiler. The SkillRuntime-Target Selector then binds each subtask to an execution method and target, producing a structured session. The \texttt{WatchdogSupervisor} claims the session and performs compatibility and safety preflight checks before constructing a \texttt{SessionRunner}. The runner coordinates the selected \texttt{SkillRuntime} and target until a terminal state is reached. Execution evidence is then written back and passed to the \texttt{SessionVerifier}, which determines whether the semantic task objective has been satisfied or whether replanning is required. Thus, planning and execution form a closed loop, but remain separated by an explicit state boundary.

\subsection{Agent Layer: From Goals to Executable Sessions}
\label{sec:agent_layer}

\begin{figure}[htbp]
    \centering
    \includegraphics[width=1\textwidth]{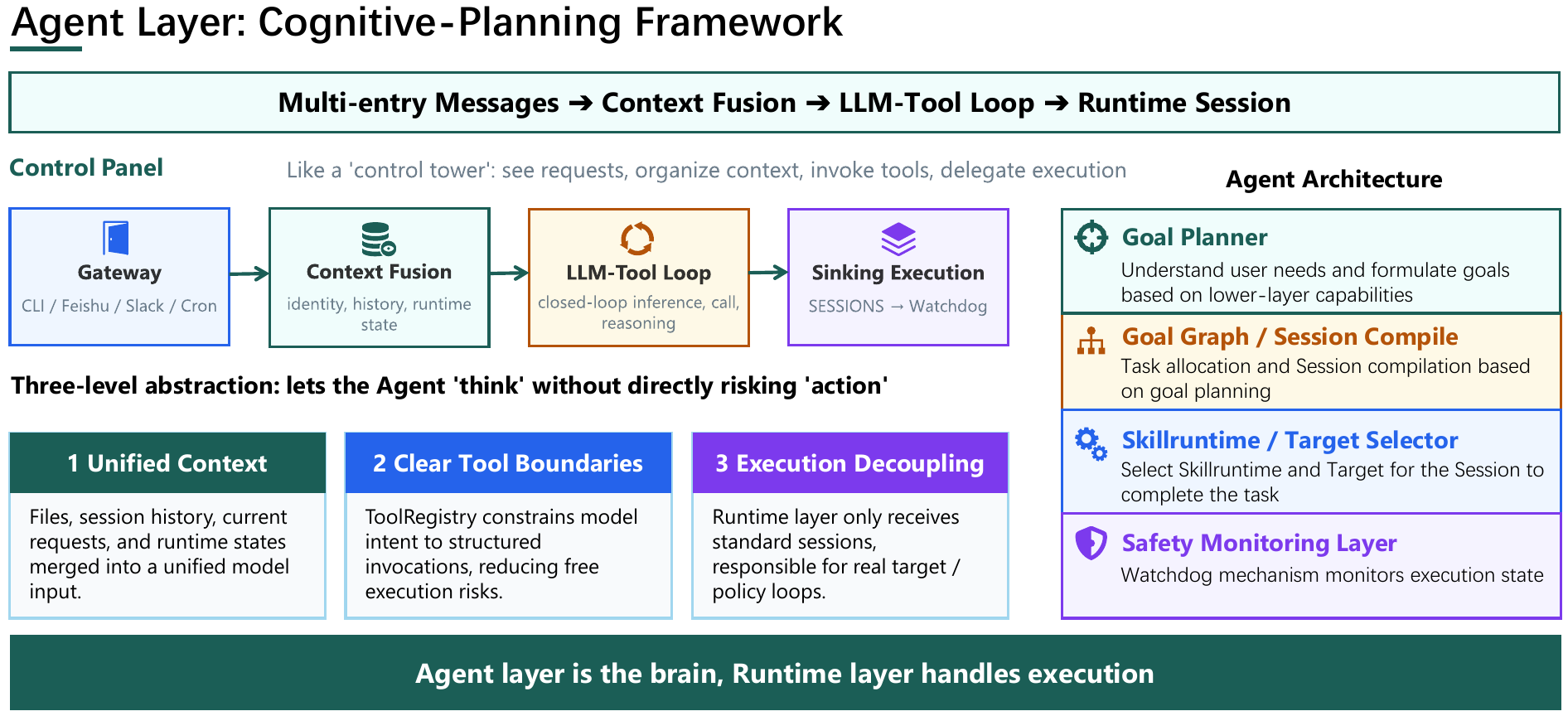}
    \caption{Agent-layer workflow. Multi-source context is converted into a goal representation, decomposed into executable sessions, and bound to a suitable \texttt{SkillRuntime} and target. The Agent produces execution contracts rather than raw target commands.}
    \label{fig:agent_layer}
\end{figure}

The Agent layer is the decision-making center of PhyAgentOS, but not a hardware controller. Its responsibility is to transform an underspecified natural-language request into a session whose execution requirements can be checked before any target is touched. Figure~\ref{fig:agent_layer} summarizes this transformation.

The planning process contains three successive commitments. First, the Goal Planner resolves the user's intent into a task objective and identifies constraints that must remain invariant during execution. Second, the Goal Graph and Session Compiler decomposes the objective into a dependency-aware subtask structure and compiles each executable unit into a session. Third, the SkillRuntime-Target Selector chooses how and where the session should run by matching the required capability, observation modalities, action semantics, and target constraints. This staged construction prevents target selection and low-level execution details from being entangled with the initial language interpretation.

Planning is grounded in the current system state rather than performed from the user request alone. The \texttt{ContextBuilder} combines the request with relevant environment state, prior attempts, available capabilities, and recovered lessons. The \texttt{ToolRegistry} exposes only structured operations for reading state, updating protocol documents, querying external resources, and initiating supported workflows. Consequently, the Agent's authority is expressed through explicit state transitions and bounded tool calls, not through unrestricted access to target SDKs.

The output of the Agent layer is a session contract containing at least the task goal, selected runtime and target, preconditions, execution limits, and acceptance criteria. After execution, the same contract anchors semantic verification: the \texttt{SessionVerifier} compares the returned evidence against the intended outcome and produces a verdict such as success, failure, or replan. The verification mechanism is discussed in Section~\ref{sec:mechanisms}; here, its architectural role is to close the cognitive loop without moving semantic judgment into the low-level controller.

\subsection{Runtime Layer: From Session Contract to Controlled Interaction}
\label{sec:runtime_layer}

Once a session has been compiled, responsibility shifts from cognitive planning to controlled execution. The Runtime layer is organized as a four-stage chain: \texttt{WatchdogSupervisor}, \texttt{SessionRunner}, \texttt{SkillRuntime}, and \texttt{Target}. Each stage narrows the scope of responsibility, from session-level governance to target-local actuation.

\begin{figure}[htbp]
    \centering
    \includegraphics[width=0.9\textwidth]{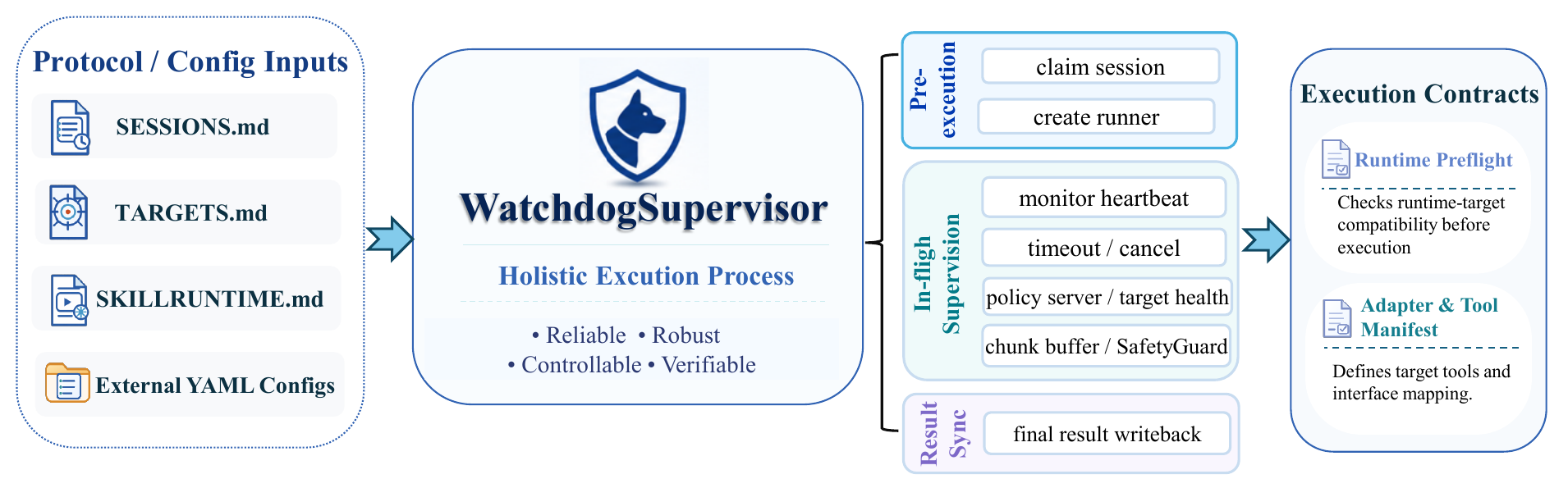}
    \caption{Session-level supervision by the \texttt{WatchdogSupervisor}. Protocol and configuration inputs are checked before execution, runner and endpoint health are monitored during execution, and terminal evidence is written back after completion.}
    \label{fig:watchdog_supervisor}
\end{figure}

The \texttt{WatchdogSupervisor} is the sole supervisory entry point. As illustrated in Figure~\ref{fig:watchdog_supervisor}, it claims pending sessions, validates their runtime contracts, creates runners, monitors heartbeats, propagates timeout or cancellation signals, and writes terminal results back to the protocol boundary. It deliberately does not perform the observation--action loop. This distinction confines scheduling and failure containment to a thin supervisory component while allowing execution strategies to evolve independently.

Before a session reaches a target, the supervisor performs a compatibility preflight. This check resolves whether the requested observation modalities, action representation, policy endpoint, target capabilities, timing constraints, and safety configuration can form a valid execution path. The result is materialized as an \texttt{AdapterPlan}, which specifies the required \texttt{PolicyAdapter}, \texttt{TargetAdapter}, and \texttt{ActionBridge} chain, together with a \texttt{TargetToolManifest} that enumerates the target operations permitted in the current session. A session that cannot produce a valid plan is rejected before target access is granted.

After preflight, the \texttt{SessionRunner} owns the lifecycle of the concrete interaction. It coordinates target resets when permitted, observation acquisition, runtime invocation, termination conditions, evidence collection, and perception refreshes. The selected \texttt{SkillRuntime} supplies the task-specific execution logic, whereas the \texttt{Target} exposes a controlled interface to the game engine, simulator process, or robot SDK. Target-local safeguards remain authoritative for constraints such as workspace bounds, speed limits, malformed commands, emergency stops, and action timing. System supervision and physical safety therefore form complementary barriers: the former governs the session, while the latter protects the embodiment at the point of actuation.

\subsection{Dual Execution Flows}
\label{sec:execution_flows}

Different embodied tasks place the decision loop at different locations. Continuous-control policies require a runtime-managed observation--inference--action loop, whereas tool-oriented tasks may require the Agent to select operations online. PhyAgentOS supports both cases through two \texttt{SkillRuntime} variants that share the same session, target, supervision, and evidence interfaces.

\subsubsection{Policy-driven execution}

In the \texttt{PolicySkillRuntime} flow, the policy is responsible for producing intelligent actions, while PhyAgentOS is responsible for making those actions executable, safe, and traceable. The Agent remains outside the low-level loop after compiling the session. The runtime repeatedly obtains an observation from the target, converts it into the model's expected input, invokes inference, converts the result into a standard action representation, and executes the resulting action or action chunk.

It can be formulated as:
\begin{equation}
    A_t = \textrm{Policy}(I, O_t, S_t, H_t)
\end{equation}
where A represents an action or action chunk, I represents a natural language instruction, O represents observation, S represents the state, and H represents history.

\begin{figure}[htbp]
    \centering
    \includegraphics[width=0.95\textwidth]{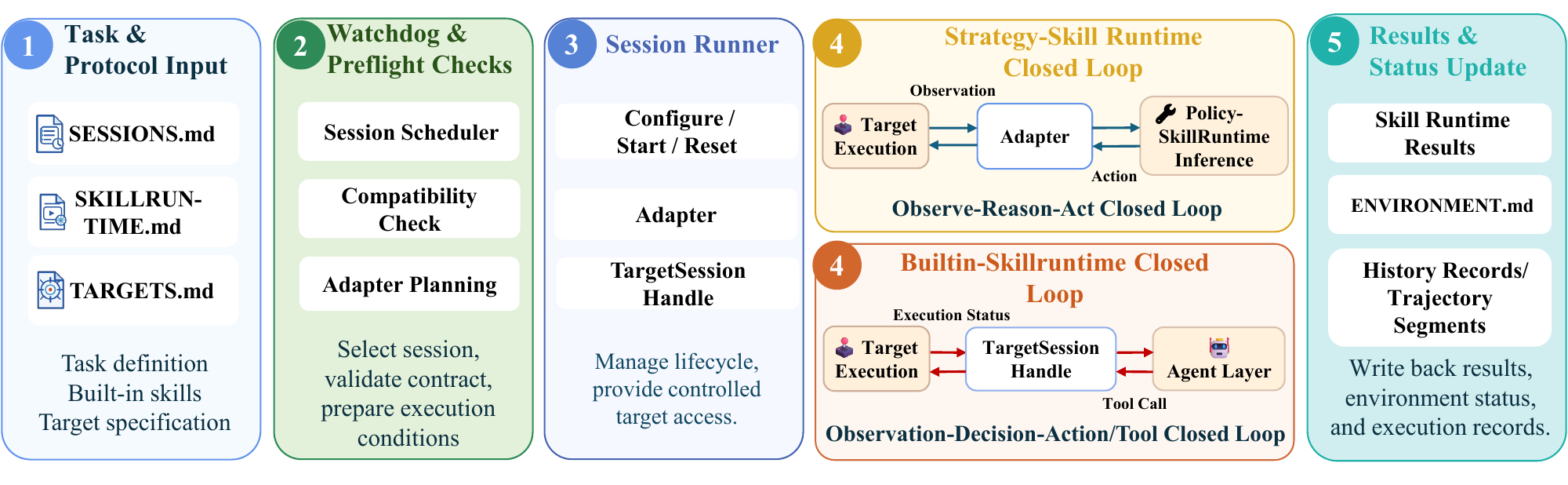}
    \caption{The SkillRuntime architecture comprises two branches. The first is the PolicySkillRuntime, designed for continuous control policies such as VLAs; it integrates models like VLAs via a unified Policy interface. In this setup, the model is responsible for generating or selecting actions, while PhyAgentOS handles session scheduling, model-to-target adaptation, safety monitoring, execution logging, and final validation. The second branch is the BuiltinSkillRuntime; it does not rely on a standalone Policy Server but instead allows the Agent or predefined logic to directly invoke controlled tools exposed by the Target. While the Agent makes step-by-step decisions, all operations remain subject to PhyAgentOS's mechanisms for permissions, safety, auditing, and verification.}
    \label{fig:policy_flow}
\end{figure}

The runtime contract intentionally does not prescribe a particular model architecture. Their heterogeneous inputs are normalized by an observation contract that can include language instructions, RGB or depth observations, robot state, and execution history. The \texttt{PolicyAdapter} converts this contract into model-specific tensors or messages and maps the model output back into one of the runtime's standard action forms, including an atomic action or an action chunk.

The Action Runtime then reconciles model time with target time. It manages action-chunk buffering, control frequency, truncation, interruption, and replanning boundaries. \texttt{ActionBridge} components enforce transformations and constraints that should remain independent of both the model and target, such as coordinate conversion, dimensional projection, clamping, and safety filtering. Finally, the \texttt{TargetAdapter} encodes the standardized action into the target-native interface. This composition allows the same model family to be evaluated across multiple targets and allows multiple model families to be compared on a common target while preserving a uniform trajectory and evidence format.

\subsubsection{Agent-directed tool execution}

In the \texttt{BuiltinSkillRuntime} flow, there is no independent policy server. Instead, the Agent participates in an online tool loop---observe, decide, invoke a tool, and observe again---through a controlled \texttt{TargetSessionHandle}. This mode is appropriate for discrete game interaction, benchmark orchestration, scripted procedures, robot primitives, and system automation, where the next operation depends on symbolic state or newly returned events rather than on a fixed continuous-control policy. It can be formulated as:
\begin{equation}
    T_t = \textrm{Agent}(I, O_t, S_t, H_t)
\end{equation}
where T represents the handle exposed by TargetSessionHandle, I represents a natural language instruction, O represents observation, S represents the state, and H represents history.


Direct participation does not imply unrestricted target access. During preflight, the \texttt{TargetToolManifest} is filtered by the session's target-tool policy, and only the resulting operations are exposed through the handle. Typical operations include \texttt{observe}, \texttt{reset}, \texttt{invoke\_tool}, \texttt{step}, and \texttt{query\_state}; dangerous or implementation-specific operations remain unavailable unless explicitly authorized. The handle validates arguments, performs adapter conversion, refreshes target state, maintains the session heartbeat, and records returned events and evidence. Thus, the Agent can dynamically adjust the execution sequence while the Runtime retains control over authority, safety, and auditability.

Despite their different decision loops, the two flows converge at the same architectural boundary. Both are created from a session, supervised by the \texttt{WatchdogSupervisor}, executed against a declared target, and terminated with a structured evidence bundle. This convergence is essential: it allows semantic verification, benchmarking, failure diagnosis, and long-term learning to operate independently of whether the underlying behavior was generated by a policy model or by an Agent-directed tool sequence.

\subsection{Protocol Layer: Explicit Contracts Across the Two Planes}
\label{sec:protocol}

\begin{figure}[htbp]
    \centering
    \includegraphics[width=1\textwidth]{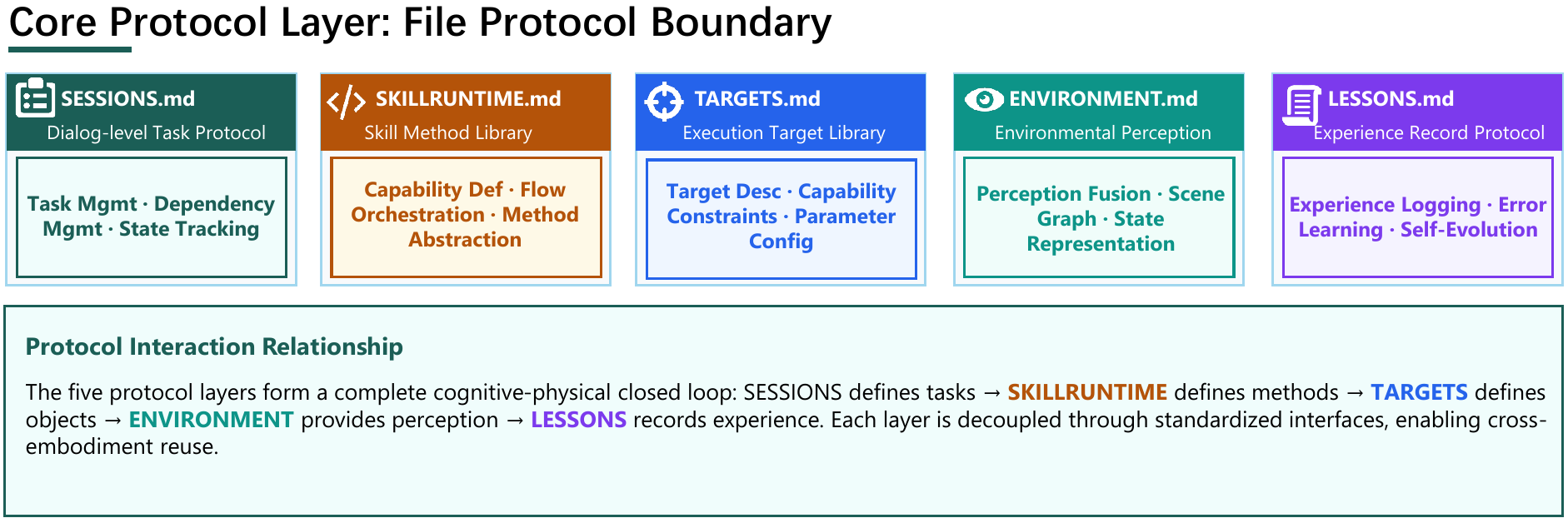}
    \caption{File-protocol boundary. Five primary Markdown documents describe the session, execution method, target, environment state, and accumulated lessons; external YAML files parameterize sensing, perception, runtime compatibility, and safety.}
    \label{fig:protocol_layer}
\end{figure}

After the responsibilities of the Agent and Runtime have been separated, the role of the protocol layer becomes precise: it is the explicit contract that connects their state transitions. The protocol is not an additional execution engine and does not encode target-specific control logic. Instead, it records what is requested, which capability and target are selected, what the system currently observes, and what was learned from the outcome. Figure~\ref{fig:protocol_layer} summarizes the five primary protocol documents.

\texttt{SESSIONS.md} is the transactional center of the architecture. Each session records a unique identifier, task objective, selected \texttt{SkillRuntime} and target, preconditions, acceptance criteria, lifecycle state, and execution result. The Agent creates or revises sessions, the supervisor atomically claims pending work, and the Runtime appends execution evidence and terminal status. The file therefore represents both the execution contract and the auditable history of how that contract was resolved.

Two documents define the two sides of execution compatibility. \texttt{SKILLRUNTIME.md} describes reusable execution methods: their required observations, produced action forms, orchestration mode, configurable parameters, and adapter requirements. \texttt{TARGETS.md} describes available execution objects: target type, capabilities, observation modalities, action semantics, endpoint information, and operational constraints. A session becomes executable only when the requirements declared by its selected runtime can be satisfied by the selected target under the current configuration.

The remaining documents close the feedback loop. \texttt{ENVIRONMENT.md} stores a structured environment snapshot derived from target observations and perception outputs, allowing the Agent and verifier to reason over entities, relations, and task-relevant state without treating raw sensor streams as the cross-layer interface. \texttt{LESSONS.md} records failure causes, corrective actions, and critic feedback so that recovery knowledge can be retrieved during future planning. Together, these documents transform a one-way command channel into a recurrent process in which execution updates the context of subsequent decisions.

The Markdown contracts are complemented by \texttt{sensors.yaml}, \texttt{perception.yaml}, \texttt{runtime\_contract.yaml}, and \texttt{safety.yaml}. These files keep deploy-time parameters separate from semantic task state: sensor schemas and synchronization rules, perception model configuration, runtime compatibility constraints, and target-specific safety limits can be changed without rewriting session logic. The resulting protocol matrix is human-readable, machine-parseable, and version-controllable, while preserving a clear distinction between task intent, execution capability, physical embodiment, observed state, and accumulated experience.

Taken together, the Agent layer, Runtime layer, dual execution flows, and protocol contracts define a complete embodied execution path. The Agent decides and compiles; the Runtime validates and acts; the protocol records the handoff and returned evidence; and semantic verification determines whether the system should accept, recover, or replan. This structure is the basis on which PhyAgentOS supports transparent execution across games, simulations, and real robots without conflating model intelligence with physical-system governance.

%% file: chapters/04_core_mechanisms/chapter.tex
\section{Core Mechanisms}
\label{sec:mechanisms}

The architecture in Section~\ref{sec:design} establishes how task intent is separated from physical execution. The remaining question is how this separation becomes a closed, self-improving system rather than a one-way pipeline from planning to action. PhyAgentOS answers this question through a sequence of mechanisms that operate on the explicit protocol state introduced in Section~\ref{sec:protocol}. First, heterogeneous observations, task dependencies, runtime status, and execution evidence are projected into a shared cognitive state space. Second, the \texttt{SessionVerifier} interprets that state to determine whether a completed execution has actually satisfied its semantic objective. Third, verification outcomes are accumulated across sessions and consolidated into retrievable knowledge, turning local trial and error into long-term adaptation. Benchmarking then scales the same loop into a controlled experimental process for measuring and validating self-evolution. A layered safety stack constrains every stage so that increased autonomy does not imply uncontrolled physical exploration.

This design is deliberate. Verification is meaningful only when task intent and physical evidence are represented in a common state space; memory is useful only when the outcomes being stored have been semantically judged; and benchmarking can support evolution only when repeated trials remain comparable, attributable, and reproducible. The following subsections therefore proceed from state alignment to semantic acceptance, from acceptance to experience consolidation, and from experience consolidation to systematic evaluation and safe deployment.

\subsection{Cognitive State Space and Dynamic Coordination}
\label{sec:cognitive_state_space}

Embodied execution is distributed not only across machines, but also across representations. The Agent reasons over goals, subtask dependencies, and symbolic entities; the Runtime observes heartbeats, adapter compatibility, and lifecycle states; policies consume tensors or structured observations; and targets expose device-specific sensor and actuator state. Directly coupling these components through private in-memory objects or pairwise RPC interfaces would leave the system without a single inspectable account of what is currently believed, what is being executed, and what evidence has been produced. PhyAgentOS instead treats the protocol layer as a shared cognitive state space: a persistent, structured projection of the task and physical world that is readable by both the cognitive and execution planes.

The state space is assembled from complementary protocol views. \texttt{SESSIONS.md} provides the transactional state of execution, including the task objective, selected runtime and target, dependency information, lifecycle status, acceptance criteria, and returned evidence. \texttt{SKILLRUNTIME.md} and \texttt{TARGETS.md} define the capability constraints under which a session can be instantiated: what observations an execution method requires, what actions it produces, and whether a candidate target can support them. \texttt{ENVIRONMENT.md} carries the task-relevant world state, whereas \texttt{LESSONS.md} extends the state space temporally by recording failure patterns and corrective knowledge recovered from previous sessions. Together, these documents make cognitive and physical state comparable without requiring either layer to import the implementation of the other.

A central operation in this alignment is the reduction of raw sensing into a structured environment representation. Target observations may contain RGB images, depth maps, point clouds, proprioceptive state, or symbolic events. The perception pipeline converts the available modalities into entities, attributes, relations, and state changes recorded in \texttt{ENVIRONMENT.md}. The Agent can therefore reason about a relation such as ``the cup is inside the cabinet'' rather than directly manipulating a pixel stream, while the Runtime can retain links to the observations from which the relation was inferred. The representation is intentionally selective: it preserves task-relevant state and evidence pointers, but does not move latency-sensitive perception or control into the protocol layer.

Task structure is aligned in the same manner. A high-level goal graph may be maintained as a directed acyclic graph in \texttt{TASK.md}, or equivalently compiled into dependency metadata attached to sessions. Nodes denote executable subtasks, and edges encode precedence or data dependencies. The compiler materializes ready nodes as pending sessions; the Runtime advances their lifecycle states; and failures invalidate only the affected node or downstream subgraph. This arrangement permits selective replanning instead of restarting an entire long-horizon task. It also supports dynamic coordination across embodiments: a navigation session and a manipulation session may be assigned to different targets while remaining synchronized through shared dependency and environment state.

The shared state space also defines the cloud--edge boundary. Resource-intensive planning, multimodal reasoning, and memory retrieval may run in the cognitive plane on remote infrastructure, whereas the \texttt{WatchdogSupervisor}, \texttt{SessionRunner}, and target-local control loop remain near the physical device. Only coarse-grained protocol state and evidence need to be synchronized across this boundary; action timing, safety interlocks, and high-frequency observations remain local. Distributed deployment therefore does not turn the network into a control bus. Instead, the network synchronizes an explicit cognitive state while the edge preserves real-time authority.

This protocol-mediated alignment is what makes the subsequent mechanisms well defined. A verifier can compare intended and observed state because both are represented in the same space. Multiple attempts can be compared because their evidence follows a common schema. Strategies can transfer across targets because task and scene semantics are separated from device-specific action encodings. The cognitive state space is therefore not merely a serialization format; it is the common reference frame in which planning, execution, verification, and learning become composable.


\subsection{SessionVerifier: Semantic Acceptance}
\label{sec:session_verifier}

A target can complete an action without completing the task that motivated it. A gripper may reach the commanded pose and close within tolerance while failing to acquire the object; conversely, an object may reach the goal region through an unplanned contact even though a nominal grasp primitive reports failure. Low-level completion signals are indispensable for control, but they describe whether an instruction was executed, not whether the intended change in the world occurred. Treating these signals as task success creates a systematic source of silent error: the execution trace terminates normally while the environment state diverges from the goal.

PhyAgentOS separates \emph{execution termination} from \emph{semantic acceptance}. After the \texttt{SessionRunner} reaches a terminal condition, the \texttt{SessionVerifier} evaluates the execution against the session contract in the shared cognitive state space. Conceptually, the verifier implements a judgment function
\[
    \mathcal{V}(G, S_0, S_T, \tau, H) \rightarrow \{\mathtt{success},\mathtt{failure},\mathtt{replan}\},
\]
where $G$ denotes the task objective and acceptance criteria, $S_0$ and $S_T$ denote the initial and terminal environment states, $\tau$ is the execution trace, and $H$ contains relevant historical context. The function need not be realized by a single model: deterministic predicates, task-specific evaluators, multimodal models, or tool-assisted review may be composed behind the same interface. What remains invariant is the evidence schema and the meaning of the returned verdict.

The evidence bundle contains the initial and terminal observations retained for the session, the original task definition, the corresponding snapshots of \texttt{ENVIRONMENT.md}, the action--observation history, and any target-native events or metrics required by the acceptance criteria. The initial state is important because success often denotes a change rather than an absolute condition; the terminal image alone cannot establish that an object was moved, a container was opened, or a hazard was avoided. By grounding judgment in both protocol state and execution evidence, the verifier evaluates the achieved world state rather than merely the controller's return code.

\begin{figure}[htbp]
    \centering
    \includegraphics[width=1\textwidth]{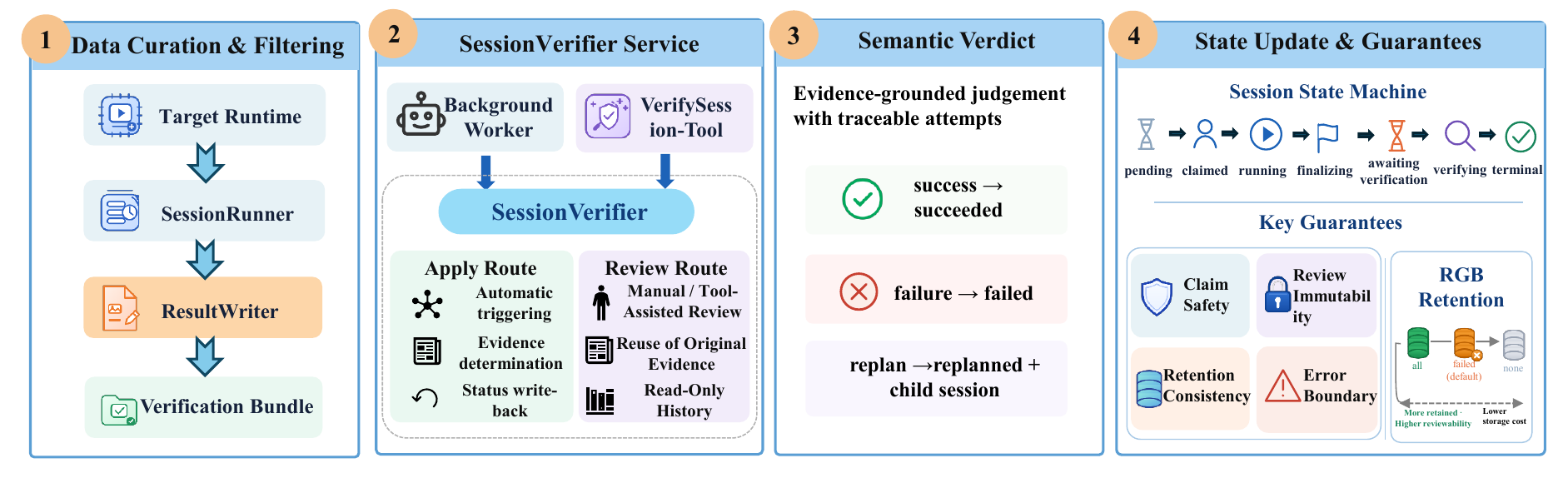}
    \caption{Semantic acceptance by the \texttt{SessionVerifier}. Runtime termination produces an evidence bundle rather than an automatic success label. The verifier compares task intent, initial and terminal state, environment snapshots, and execution history to return \texttt{success}, \texttt{failure}, or \texttt{replan}.}
    \label{fig:session_verifier}
\end{figure}

The three verdicts induce different state transitions. \texttt{success} indicates that the evidence satisfies the acceptance criteria and moves the session to \texttt{succeeded}. \texttt{failure} indicates that the objective was not achieved and moves the session to \texttt{failed}; the evidence is then available for diagnosis and lesson extraction. \texttt{replan} indicates that the current evidence supports a revised continuation rather than unconditional acceptance or rejection. In this case, the original attempt remains immutable and a child session is compiled with updated preconditions, target state, or strategy. Replanning therefore creates a new auditable decision rather than rewriting the history of the previous execution.

Every judgment is appended to the session's \texttt{attempts} record together with the verifier configuration and evidence references used to produce it. Automatic acceptance and tool-mediated review share this append-only interface, so later inspection does not erase earlier outcomes. A session that fails on its first attempt and succeeds after replanning retains both verdicts, their evidence bundles, and the causal relation between the parent and child sessions. Storage policy may determine whether raw RGB evidence is retained indefinitely, but the judgment, environment snapshot, provenance, and terminal state remain available for audit.

Semantic acceptance closes the immediate execution loop, but a verdict alone does not make the system improve. Without consolidation, the next similar session would begin with no knowledge of why the previous attempt failed or which recovery succeeded. The verifier therefore supplies the labeled evidence on which long-term trial and error is built.

\subsection{Closed-loop Trial-and-Error and Experience Consolidation}
\label{sec:epistemic_memory}

Self-evolution in PhyAgentOS is not defined as unconstrained online modification of a policy model. It is first a system-level process by which verified execution outcomes alter the context, strategy selection, and reusable capabilities available to future sessions. The \texttt{SessionVerifier} provides the critical supervision signal: it converts a session from an unlabeled trajectory into an accepted success, an attributed failure, or a continuation requiring replanning. Epistemic Memory extends this signal across time.

The process forms a closed loop with six stages. \emph{Execute}: a session is run through the standard Runtime path and produces a trace and terminal evidence. \emph{Verify}: the \texttt{SessionVerifier} assigns a semantic verdict. \emph{Diagnose}: for a failure or replan verdict, the Agent identifies a candidate cause by jointly examining the task contract, environment transition, runtime events, and prior lessons. \emph{Revise}: a recovery strategy modifies the subgoal, selected runtime, target configuration, or action method and is compiled into a new session. \emph{Re-verify}: the revised strategy is executed under the same acceptance semantics. \emph{Consolidate}: only after the outcome is verified is the resulting knowledge committed to persistent memory. This ordering prevents speculative explanations and untested corrections from being treated as established experience.

Experience is organized at three levels of abstraction. The \emph{episodic level} preserves individual sessions, including their task, trace, evidence, verdict, and parent--child relations. These records support replay and forensic analysis. The \emph{semantic level} aggregates repeated episodes into task- and environment-level knowledge, such as stable object relations, recurrent failure modes, or conditions under which one strategy outperforms another. The \emph{methodological level} captures procedures that remain effective across sufficiently diverse initial states. Such procedures can be promoted into reusable skill specifications and registered through \texttt{SKILLRUNTIME.md}, allowing future planning to retrieve a validated method rather than synthesize it from scratch.

Two persistent knowledge interfaces distinguish positive and corrective experience. \texttt{KNOWLEDGE.md} stores validated successful patterns together with their applicability conditions, target and runtime provenance, and observed performance. \texttt{LESSONS.md} stores failure-centered records: the failed objective, the relevant evidence, the diagnosed cause, the attempted correction, and whether that correction was subsequently verified. This separation avoids the common failure mode in which a recovery hypothesis is remembered as fact merely because it was proposed. An entry becomes a reusable lesson only when its corrective effect is supported by a later verdict.

\begin{figure}[htbp]
    \centering
    \includegraphics[width=1\textwidth]{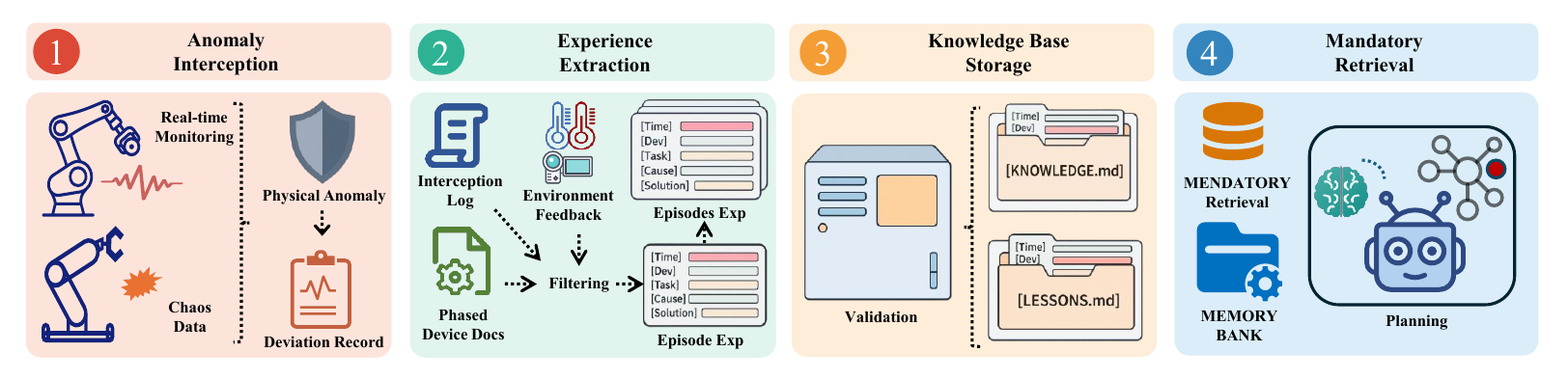}
    \caption{Closed-loop trial and error. Session evidence is semantically verified, diagnosed, revised, re-executed, and consolidated into episodic records, semantic knowledge, and reusable methods. Successful practices are stored in \texttt{KNOWLEDGE.md}; verified failure corrections are stored in \texttt{LESSONS.md}.}
    \label{fig:epistemic_memory}
\end{figure}

Retrieval closes the long-term loop. Before compiling a new session, the \texttt{ContextBuilder} queries memory using the current goal, environment state, target type, and detected risk factors. Retrieved knowledge is injected with provenance and scope: a lesson learned on one embodiment is not assumed to transfer unless its preconditions match the new target and scene. This conditional retrieval is essential for cross-target reuse. Symbolic knowledge such as task ordering may transfer broadly, whereas a contact-rich manipulation strategy may remain specific to a gripper geometry or control frequency.

The result is cumulative adaptation without requiring every improvement to be encoded through neural retraining. The system can become more capable by selecting better methods, avoiding previously observed failures, refining acceptance conditions, and promoting repeatedly verified procedures. Model fine-tuning may be added as a further consolidation mechanism, but the protocol-level loop already provides a measurable form of self-evolution: later sessions are informed by verified outcomes from earlier ones.

Individual improvements, however, can be anecdotal. A strategy that succeeds after one retry may exploit a favorable initial state, and a growing memory may improve some task categories while degrading others. PhyAgentOS therefore uses benchmarking to turn repeated verified sessions into controlled evidence about whether the system is genuinely evolving.

\subsection{Benchmarking as an Instrument for Self-Evolution}
\label{sec:benchmarking}

Benchmarking in PhyAgentOS is not a separate script wrapped around a policy. It is a controlled orchestration mode built on the same session, runtime, verification, and memory mechanisms used during ordinary operation. This design serves two purposes. It removes discrepancies between development and evaluation paths, and it provides the repeated, structured trials required to distinguish reliable improvement from isolated success.

\begin{figure}[htbp]
    \centering
    \includegraphics[width=1\textwidth]{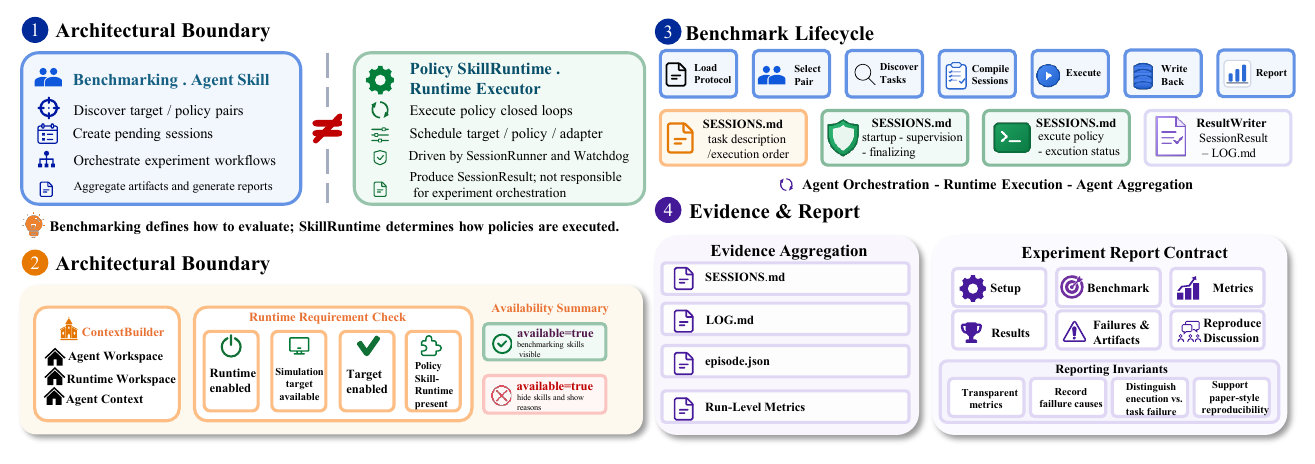}
    \caption{Benchmarking as runtime orchestration. An availability gate validates the required simulation target and policy runtime, a compiler materializes task--initial-state pairs as sessions, and the standard Watchdog--Runner--SkillRuntime path executes and verifies every episode.}
    \label{fig:benchmarking}
\end{figure}

A benchmark run begins with an \textbf{Availability Gate}. The gate verifies that the required simulation target is enabled and reachable, that the selected policy or built-in runtime can be instantiated, and that the corresponding adapter, observation, action, and safety contracts pass preflight. A failed gate produces an explicit diagnostic rather than a partial result set whose missing episodes are silently ignored.

The \textbf{Session Compiler} then expands benchmark tasks and initial conditions into a reproducible set of pending sessions. Each session records the task source, initial-state identifier or random seed, selected runtime and target, evaluation configuration, and acceptance criteria. The \texttt{WatchdogSupervisor} claims these sessions and dispatches them through the same \texttt{SessionRunner}, \texttt{SkillRuntime}, adapter chain, and target interface used elsewhere in the system. Episode artifacts, traces, verifier attempts, and terminal results are written back through the standard protocol boundary rather than through benchmark-specific logging code.

This reuse is methodologically important. When a policy is evaluated, observation normalization, action conversion, timeouts, chunk execution, and semantic acceptance are identical to those used in deployment. Reported differences can therefore be attributed to declared experimental variables rather than to divergent harness implementations. Each aggregate metric remains traceable to the session identifiers and artifact paths from which it was computed.

The reporting layer aggregates both outcome and diagnostic measures. Typical measures include Success Rate, per-task Success Rate, action count or episode length, task-specific reward, and an error taxonomy separating perception, policy, adapter, execution, timeout, safety, and verifier-related failures. Because failure episodes already produce structured evidence and lesson candidates, the taxonomy can be derived from the same causal records used by the self-evolution loop instead of being reconstructed manually after evaluation.

Benchmarking also separates \emph{measurement} from \emph{adaptation}. A frozen evaluation pass measures the current system under a fixed configuration. Verified failures may then be used to update \texttt{LESSONS.md}, revise orchestration, or promote a new method. The revised system is evaluated on a new or held-out session set, with configuration and memory versions recorded. Agent-assisted retry can be reported as a distinct protocol, but it is not conflated with first-attempt policy performance. This separation preserves scientific validity while still allowing the benchmark to act as a data engine for iterative improvement.

In this formulation, benchmarking can further validate epistemic memory. The memory mechanism proposes that accumulated verified experience improves future behavior; benchmarking tests that proposition at scale, localizes where gains occur, and detects regressions. The same mechanism that generates an evaluation table also produces a reproducible corpus of successful trajectories, failed attempts, recovery sessions, and causal diagnoses for further evolution.

\subsection{Safety Mechanisms}

The preceding mechanisms allow PhyAgentOS to execute, verify, retry, and improve physical behavior. Such autonomy is meaningful only when every execution remains within an explicit safety envelope. PhyAgentOS therefore adopts a defense-in-depth design composed of five complementary mechanisms: compatibility preflight, action bridges, \texttt{SafetyGuard}, heartbeat monitoring, and target-local constraints. These mechanisms operate at different stages of the execution path and address distinct failure classes, ranging from incompatible component composition to unsafe commands, communication failures, and device-level hazards.

\textbf{Compatibility preflight.}
Safety begins before a session is admitted for execution. The \texttt{WatchdogSupervisor} evaluates whether the selected \texttt{SkillRuntime}, policy, adapters, and target form a valid execution chain. The preflight procedure checks required observation modalities, action schemas, coordinate conventions, control frequencies, endpoint availability, and the operations exposed by the target. It also verifies that the session's requested capabilities are permitted by the corresponding tool and safety configurations. The resulting \texttt{AdapterPlan} and \texttt{TargetToolManifest} make these assumptions explicit. A session is rejected before target access is created when its required modality is unavailable, its action representation is ambiguous, or its requested operation exceeds the declared authority boundary. Compatibility preflight thus prevents structurally invalid or unauthorized configurations from entering the physical execution loop.

\textbf{Action bridges.}
Once a session is admitted, action bridges provide a deterministic boundary between runtime-level action representations and target-specific commands. They perform declared transformations such as coordinate-frame conversion, unit normalization, joint reordering, dimensional projection, gripper remapping, and action-chunk resampling. By isolating these transformations from both the policy and the target, PhyAgentOS avoids embedding platform-specific assumptions inside reusable skills. Each bridge is selected and type-checked during preflight, making the complete action-conversion path inspectable before execution. Bounded transformations may also project values into a predefined valid range when such projection preserves the intended semantics; malformed, underspecified, or discontinuously unsafe commands are rejected rather than silently repaired. Action bridges therefore provide adaptation without turning format conversion into an implicit source of uncontrolled behavior.

\textbf{\texttt{SafetyGuard}.}
After conversion into the target action space, every action or action chunk is examined by \texttt{SafetyGuard} before transmission. In contrast to the action bridge, which defines how an action is represented, \texttt{SafetyGuard} determines whether the resulting command is admissible. Its target-configured checks include data-type and dimensional validation, rejection of NaN and infinite values, joint and workspace limits, velocity and acceleration bounds, command duration, action frequency, and emergency-stop state. Depending on the violated constraint, the command may be rejected, replaced by a safe halt, or bounded through an explicitly authorized transformation. Every intervention is recorded with a structured violation code and attached to the session evidence. Consequently, a safety intervention remains distinguishable from a policy error, communication failure, or ordinary task failure during subsequent semantic verification.

\textbf{Heartbeat monitoring.}
Command-level validation cannot detect failures caused by stalled processes, network partitions, or disconnected policy and target services. The \texttt{WatchdogSupervisor} therefore monitors heartbeats from the \texttt{SessionRunner}, policy service, and target runtime throughout the session. Missing or inconsistent heartbeats trigger timeout handling, cancellation propagation, and controlled termination rather than allowing execution to continue under stale state. Heartbeat monitoring also bounds the lifetime of outstanding action chunks and prevents a component from retaining target authority after its session has become invalid. This mechanism provides session-level fault containment: it identifies which component became unavailable, terminates the affected execution path, and writes an attributed failure state for later verification and diagnosis.

\textbf{Target-local constraints.}
The target remains the final authority over what may physically occur. Real robots retain local joint limits, collision detection, torque or velocity constraints, controller watchdogs, workspace restrictions, and hardware emergency stops Simulators enforce their own action validity, reset, collision, and workspace policies, while game targets restrict privileged commands and invalid state transitions. These constraints are enforced close to the actuator or environment state and remain active even when an upstream runtime, bridge, or communication channel fails. PhyAgentOS does not replace such native safeguards; it treats them as the innermost and most authoritative layer of the safety envelope.

Together, the five mechanisms establish a progressively stronger safety boundary. Compatibility preflight constrains which component compositions may execute; action bridges constrain how abstract actions are translated; \texttt{SafetyGuard} constrains which translated commands may be issued; heartbeat monitoring constrains execution under component and communication failures; and target-local constraints bound the final physical effect. Safety events are written into the shared cognitive state and included in the evidence bundle, allowing the \texttt{SessionVerifier} to distinguish unsafe goal pursuit from safety-preserving termination. They may also be consolidated into \texttt{LESSONS.md} as reusable failure knowledge, while the underlying constraints remain immutable to the self-evolution process. Safety thus defines the admissible region within which PhyAgentOS can explore, recover, and improve.

%% file: chapters/05_experiments/chapter.tex
\section{Experiments}
\label{sec:experiments}

Evaluating embodied AI systems involves a persistent tension. Real robots provide the most authentic test conditions, yet experiments are slow, hardware is scarce, and failures carry physical risk. Simulators offer speed and safety, but they introduce artifacts that make it hard to tell whether a result reflects genuine intelligence or merely simulator-specific dynamics. We resolve this tension through progressive validation. Cognitive strategies are first tested under idealized conditions where intelligence is the only variable. Physical constraints are then reintroduced gradually, first through simulation and finally on real hardware. The same protocol files and runtime infrastructure are used at every tier. Any performance drop can therefore be traced to a specific layer rather than to an architectural collapse.

The evaluation is organized into three tiers. Section~\ref{sec:game_agent} evaluates cognitive capabilities in controlled game environments. Section~\ref{sec:simulation} introduces rigid-body dynamics, collision, and control latency through simulation. Section~\ref{sec:real_robot} evaluates safety and cross-embodiment deployment on real hardware.

\subsection{Game-Tier Cognitive Verification}

\label{sec:game_agent}

\begin{figure}[htbp]
    \centering
    \includegraphics[width=1\textwidth]{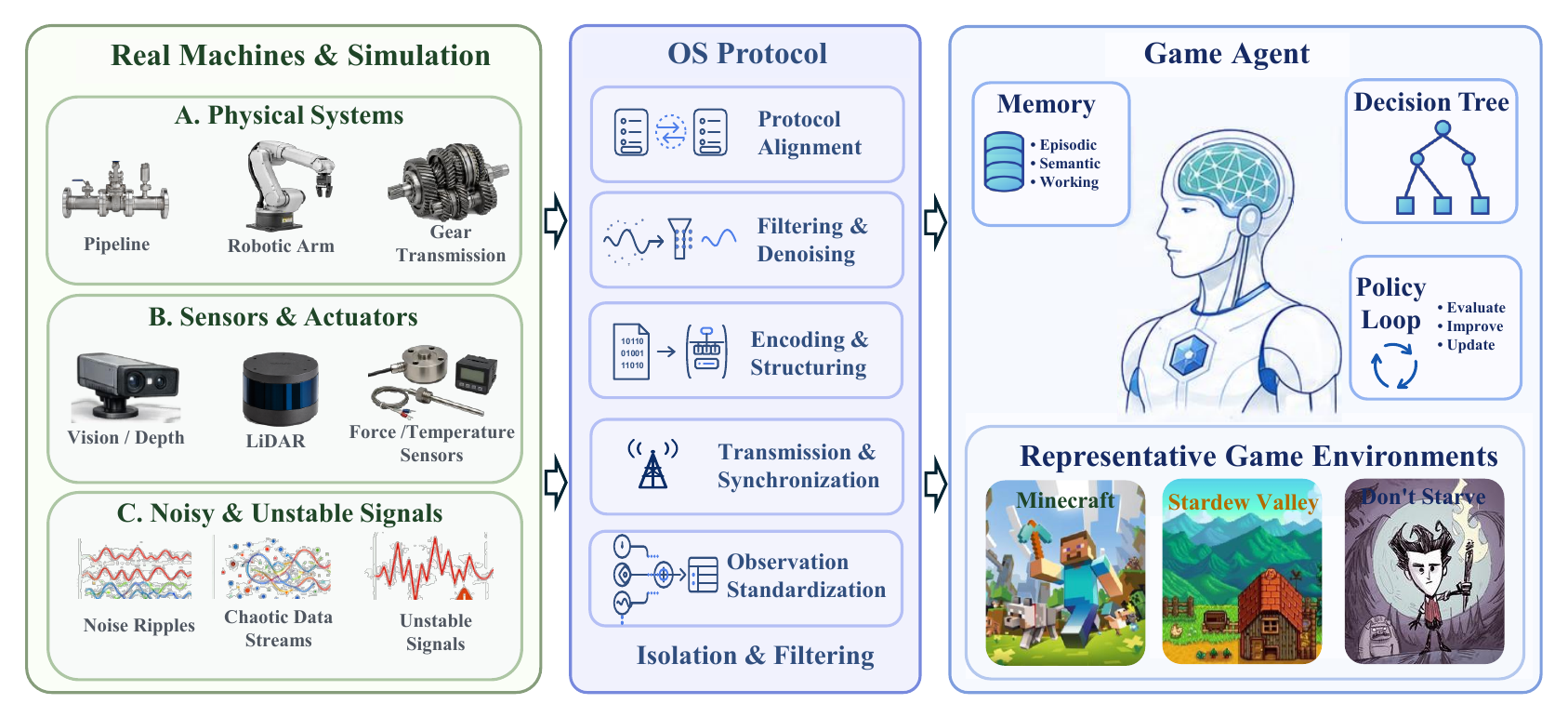}
    \caption{General Game Agent as a controlled-variable paradigm. Game environments strip away physical assumptions, isolating cognition as the object of study. The progressive environment ladder ranges from resource-constrained Minecraft to multi-dimensional survival in Don't Starve.}
    \label{fig:game_agent}
\end{figure}

The game tier is designed to test cognition in isolation. In a game engine, actions are executed faithfully. A grasp command always succeeds. A navigation command never drifts. The visual observation is never corrupted by sensor noise. Any failure to achieve the task must therefore be located in the cognitive layer, not in low-level control. This controlled-variable paradigm lets us study memory, planning, and self-evolution as pure decision-making problems.

Three environments are chosen to stress different cognitive dimensions, as shown in Figure~\ref{fig:game_agent}. Minecraft tests goal pursuit under sparse feedback, where the agent must gather materials and craft tools. Stardew Valley tests multi-resource scheduling, where the agent must manage seasonal crop cycles, trade in an in-game economy, and maintain social relations while monitoring energy, money, and time. Don't Starve tests risk-aware planning under existential threat, where the agent faces irreversible permadeath, seasonal hazards, and a deep technology tree that gates access to advanced tools. Together, these environments span a spectrum from structured task completion to open-ended survival.

Across all three environments, three cognitive capabilities are evaluated. Memory is measured by whether the agent retains knowledge across sessions. We test retrieval accuracy when a situation recurs, generalization to novel regions, and retention over tens or hundreds of sessions. Risk is measured by resource management under survival pressure. We ask whether the agent allocates limited energy and time to actions that maximize long-term survival rather than immediate reward. Self-evolution is measured by the closure of the fail-reflect-improve-reverify loop. When the agent encounters a novel failure, does it produce a recovery strategy, test it, and consolidate the lesson so that the same failure does not recur?

\subsubsection{Experimental Setup}

All game experiments are conducted through the standard PhyAgentOS pipeline. The \texttt{WatchdogSupervisor} claims pending sessions from \texttt{SESSIONS.md}. The \texttt{SessionRunner} drives the game target through the appropriate \texttt{TargetAdapter}. The \texttt{SessionVerifier} evaluates outcomes against the acceptance criteria defined in each session contract. Epistemic Memory is active throughout. Successful strategies are promoted to \texttt{SKILL.md}. Failures are consolidated into \texttt{LESSONS.md}. The cognitive layer is held constant across all three environments. Only the target configuration and the corresponding \texttt{TargetAdapter} are changed to match each game engine.

\input{chapters/05_experiments/table/mc}
For Minecraft, the upper layer uses qwen3.5-plus for task planning and qwen3-vl-flash for visual perception. The lower layer uses STEVE-1 \cite{lifshitz2023steve1} to translate subgoals into keyboard and mouse actions. For Stardew Valley, the agent operates through a text-only interface with deepseek-v4-flash as the reasoning backbone. The \texttt{SessionVerifier} provides the semantic acceptance signal for each Lite-100 \cite{tan2025stardojo} task. For Don't Starve, the same text-only configuration is used. \texttt{ENVIRONMENT.md} captures survival-relevant state variables at each observation step. LESSONS.md records death causes to inform recovery strategies on the next rebirth.

\subsubsection{Results on Optimus-67}

Optimus-67 contains 67 long-horizon Minecraft tasks organized into seven groups of increasing difficulty. These are Wood, Stone, Iron, Gold, Diamond, RedStone, and Armor. The tasks require sustained resource acquisition, mining, crafting, and equipment production. They test an agent's capacity for long-horizon planning and execution under sparse feedback.

The results are shown in Table \ref{tab:optimus3}. PhyAgentOS achieves near-saturated performance on Wood and Stone, with success rates of 99\% and 96\%. On Iron, the rate drops to 52\%. This is comparable to the strongest baseline and reflects the substantial increase in planning complexity needed to locate and smelt ore. Gold and Diamond remain challenging for all methods. PhyAgentOS achieves 6\% and 19\%, respectively. The highest relative advantage is found on RedStone, where PhyAgentOS reaches 30\% and exceeds all reported baselines. This pattern is consistent with two mechanisms. First, the SessionVerifier can detect partial circuit constructions that binary completion metrics would count as failures. Second, Epistemic Memory retains and reuses redstone recipes across sessions. On Armor, the rate of 15\% is competitive but indicates that end-game equipment production remains a frontier task.

\subsubsection{Results on StarDojo}

\input{chapters/05_experiments/table/star}

StarDojo Lite-100 provides 100 procedurally varied tasks that require sustained goal-directed behavior under tight token and latency budgets. The benchmark reports standard Success Rate, Budgeted Success Rate, and efficiency metrics. The PhyAgentOS run used the text-only model deepseek-v4-flash and achieved 22.0\% overall success. This exceeds the SPIKE baseline of 18.0\%. For reference, GPT-4.1 achieves 12.7\% on the same split in a multimodal configuration. The PhyAgentOS run is text-only.

Success is concentrated on easy tasks. The rate is 37.5\% on 56 easy tasks. Medium and hard tasks remain challenging at 3.7\% and 0.0\%. The mean progress rate is 0.236. Solved tasks average 12.5 steps. Most episodes end in failure. The remainder is split into success, truncation, and a single knockout. The breakdown by task category is shown in Table \ref{tab:star_dojo}. The largest improvement over baselines appears in Crafting. PhyAgentOS reaches 50.0\% against 23.8\% for the strongest baseline. This gain is attributed to two factors. First, validated recipes accumulate in KNOWLEDGE.md. Second, the SessionVerifier accepts partially completed craft sequences that reactive baselines reject.

\subsubsection{Results on DST-Dojo}

\input{chapters/05_experiments/table/dst}

DST-Dojo evaluates survival competence through four axes. Survival efficiency measures days survived at each rebirth count. It reveals whether the agent learns from death or repeats the same mistakes. Seasonal robustness disaggregates survival by season. It tests preparation for freezing, overheating, and rain-induced sanity loss. Tech tree progression measures the day at which each major research station is unlocked. Resource management tracks hunger and sanity as continuous signals of survival competence.

The pilot results for 10 episodes in Autumn, Day 0, Day Phase are shown in Table \ref{tab:dst_dojo}. PhyAgentOS doubles mean survival from 1.02 days to 2.10. It achieves a 30\% Day 3 survival rate where the baseline achieves none. Death by darkness drops from 90\% to 80\%. This indicates that LESSONS.md is partially effective at preventing repeated fatal errors, though darkness remains the dominant failure mode. Death by starvation rises from 0\% to 10\%. This reflects the agent's extended lifespan and eventual food depletion. The lower average health, hunger, and sanity values in the PhyAgentOS condition are not signs of degradation. They reflect sustained resource consumption over a longer survival period.

\subsubsection{Cross-Environment Cognitive Analysis}

The results across the three environments provide convergent evidence that the cognitive layer functions effectively when physical execution is abstracted. Memory is demonstrated by the retention of crafting recipes in Minecraft and Stardew Valley, and by the reduction in repeated death causes in Don't Starve. Planning is demonstrated by progression through increasingly complex task groups in Optimus-67 and by multi-step craft sequences in StarDojo. Self-evolution is demonstrated by the SessionVerifier-driven retry loop and the consolidation of failures into persistent memory. This produces measurable improvements in later sessions.

Yet these results do not establish that the same strategies will survive when physical execution is no longer abstracted. The game tier validates the cognitive layer in isolation. The critical question is whether these strategies degrade gracefully or collapse entirely under physical constraints. That question is addressed in the next section.

\subsection{Simulation-Tier Physical Regression}
\label{sec:simulation}

Simulation provides a controlled physical validation layer between game-based cognitive evaluation and real-robot deployment. We evaluate PhyAgentOS on three simulation benchmarks that span different levels of embodied difficulty: LIBERO for suite-level language-conditioned tabletop manipulation, CALVIN for long-horizon chained manipulation under persistent physical state, and RoboCasa365 for household activity execution in high-fidelity kitchen environments. Across all benchmarks, we use a dual evaluation protocol. The \emph{First} setting measures the policy's original first-attempt performance. The \emph{Final} setting measures performance after PhyAgentOS intervention, where the verifier is invoked only after failure and may trigger a controlled recovery attempt from the current physical state. The policy weights, nominal task goals, and benchmark success criteria are left unchanged.

\subsubsection{LIBERO Manipulation Benchmark}

LIBERO is a language-conditioned manipulation benchmark covering four task suites: Spatial, Object, Goal, and LIBERO-10. These suites evaluate spatial reasoning, object-centric manipulation, goal-conditioned execution, and long-horizon manipulation, respectively.

We evaluate four policy backends: OpenVLA~\cite{kim2024openvla}, $\pi_0$~\cite{black2025pi0}, $\pi_{0.5}$~\cite{intelligence2025pi05}, and X-VLA~\cite{zheng2025xvla}. For each policy, PhyAgentOS is allowed to intervene only after an unsuccessful first attempt. The recovery process does not fine-tune the policy or rewrite the benchmark task; it uses verifier feedback to identify recoverable execution failures and resume control.

\begin{table}[htbp]
\centering
\caption{Agent-assisted LIBERO validation. Values are success rates. ``Final'' denotes performance after verifier-triggered retry.}
\label{tab:libero_agent_assisted}
\small
\setlength{\tabcolsep}{4pt}
\begin{tabular}{llccccc}
\toprule
\textbf{Model} & \textbf{Setting} & \textbf{Spatial} & \textbf{Object} & \textbf{Goal} & \textbf{LIBERO-10} & \textbf{Overall} \\
\midrule
\multirow{3}{*}{OpenVLA}
& First & 84.4\% & 86.4\% & 74.6\% & 52.6\% & 74.5\% \\
& Final & 85.8\% & 87.6\% & 75.2\% & 53.4\% & 75.5\% \\
& Gain & +1.4 & +1.2 & +0.6 & +0.8 & +1.0 \\
\midrule
\multirow{3}{*}{$\pi_0$}
& First & 96.8\% & 98.2\% & 93.4\% & 82.6\% & 92.8\% \\
& Final & 97.6\% & 98.2\% & 94.2\% & 82.6\% & 93.2\% \\
& Gain & +0.8 & +0.0 & +0.8 & +0.0 & +0.4 \\
\midrule
\multirow{3}{*}{$\pi_{0.5}$}
& First & 99.4\% & 98.8\% & 97.2\% & 92.8\% & 97.0\% \\
& Final & 99.6\% & 98.8\% & 98.0\% & 94.6\% & 97.8\% \\
& Gain & +0.2 & +0.0 & +0.8 & +1.8 & +0.8 \\
\midrule
\multirow{3}{*}{X-VLA}
& First & 97.2\% & 97.4\% & 98.0\% & 96.6\% & 97.3\% \\
& Final & 99.0\% & 97.8\% & 99.0\% & 98.6\% & 98.6\% \\
& Gain & +1.8 & +0.4 & +1.0 & +2.0 & +1.3 \\
\bottomrule
\end{tabular}
\end{table}

Table~\ref{tab:libero_agent_assisted} shows consistent improvements across all four policy backends. Overall success increases by +1.0 point for OpenVLA, +0.4 points for $\pi_0$, +0.8 points for $\pi_{0.5}$, and +1.3 points for X-VLA. The gains are modest because several LIBERO policies already operate at high first-attempt success rates, but PhyAgentOS still recovers residual execution failures, including on the more demanding LIBERO-10 suite.

\subsubsection{CALVIN Long-Horizon Manipulation Benchmark}

CALVIN evaluates language-conditioned long-horizon manipulation in a simulated tabletop environment. We use the standard ABC$\rightarrow$D generalization setting, where policies are tested on unseen environment configurations. Each episode consists of a five-subtask language sequence that must be completed without resetting the environment between subtasks. This protocol stresses persistent physical state, compounding execution errors, and mid-sequence recovery.

We evaluate three policy backends on CALVIN: $\pi_0$~\cite{black2025pi0}, $\pi_{0.5}$~\cite{intelligence2025pi05}, and X-VLA~\cite{zheng2025xvla}. In the First setting, each subtask is attempted once by the policy. In the Final setting, PhyAgentOS invokes the verifier after a failed subtask, diagnoses the failure, and, when appropriate, generates a recovery instruction before resuming from the current state. We report the average number of successfully completed subtasks per five-step sequence (Avg. Len.) and the chain success rate after each subtask. A $k/5$ success rate denotes the fraction of sequences that complete at least the first $k$ subtasks.

\begin{table}[htbp]
\centering
\caption{Agent-assisted CALVIN ABC$\rightarrow$D validation. Values are chain success rates unless otherwise noted. ``Final'' denotes performance after verifier-triggered recovery. Avg. Len. is the average number of completed subtasks per five-subtask sequence.}
\label{tab:calvin_agent_assisted}
\small
\setlength{\tabcolsep}{4pt}
\begin{tabular}{llcccccc}
\toprule
\textbf{Model} & \textbf{Setting} & \textbf{1/5} & \textbf{2/5} & \textbf{3/5} & \textbf{4/5} & \textbf{5/5} & \textbf{Avg. Len.} \\
\midrule
\multirow{3}{*}{X-VLA}
& First & 96.8\% & 92.0\% & 86.2\% & 81.7\% & 74.3\% & 4.310 \\
& Final & 97.0\% & 92.0\% & 86.3\% & 81.9\% & 75.7\% & 4.329 \\
& Gain & +0.2 & +0.0 & +0.1 & +0.2 & +1.4 & +0.019 \\
\midrule
\multirow{3}{*}{$\pi_0$}
& First & 86.5\% & 74.0\% & 62.2\% & 50.9\% & 38.9\% & 3.125 \\
& Final & 86.8\% & 74.7\% & 64.1\% & 53.7\% & 45.6\% & 3.249 \\
& Gain & +0.3 & +0.7 & +1.9 & +2.8 & +6.7 & +0.124 \\
\midrule
\multirow{3}{*}{$\pi_{0.5}$}
& First & 99.7\% & 98.0\% & 94.5\% & 91.5\% & 85.3\% & 4.690 \\
& Final & 99.7\% & 98.5\% & 95.2\% & 92.5\% & 89.4\% & 4.753 \\
& Gain & +0.0 & +0.5 & +0.7 & +1.0 & +4.1 & +0.063 \\
\bottomrule
\end{tabular}
\end{table}

Table~\ref{tab:calvin_agent_assisted} shows that PhyAgentOS improves long-horizon execution for all three CALVIN backends. The improvement is most visible in full-chain completion: X-VLA improves from 74.3\% to 75.7\%, $\pi_0$ from 38.9\% to 45.6\%, and $\pi_{0.5}$ from 85.3\% to 89.4\%. Average completed subtasks also increase for every model. These gains indicate that a portion of CALVIN failures are recoverable mid-sequence execution errors rather than irreversible policy incapabilities.

\subsubsection{RoboCasa365 Household Manipulation Benchmark}

RoboCasa365 evaluates household manipulation in high-fidelity kitchen environments~\cite{robocasa365}. Compared with tabletop-only manipulation suites, it places greater emphasis on cluttered indoor scenes, articulated objects, multi-view perception, and long-horizon household activities. We evaluate the official \texttt{target50} task set under the leaderboard-aligned \texttt{pretrain} split, which contains 18 atomic skills and 32 composite activities. Each task is evaluated with five randomized trials, yielding 250 episodes in total. Success is measured by the environment completion signal.

We evaluate three policy backends on this benchmark: $\pi_{0.5}$~\cite{intelligence2025pi05}, RLDX-1~\cite{rldx2026}, and WorldDreamer~\cite{worlddreamer2026robocasa}. In the First setting, the policy executes one rollout from the initial environment reset. In the Final setting, PhyAgentOS intervenes only after an episode-level failure. The verifier may either overturn a false-negative completion judgment or produce a recovery instruction that allows execution to continue from the current physical state. The environment is not reset during recovery, the task goal is not changed, and the policy weights remain fixed.

\begin{table}[htbp]
\centering
\caption{Agent-assisted RoboCasa365 \texttt{target50} validation on the \texttt{pretrain} split. Values are episode success rates. Atomic contains 18 skills / 90 episodes, and Composite contains 32 activities / 160 episodes. ``Final'' denotes performance after verifier-triggered recovery.}
\label{tab:robocasa365_agent_assisted}
\small
\setlength{\tabcolsep}{5pt}
\begin{tabular}{llcccc}
\toprule
\textbf{Model} & \textbf{Setting} & \textbf{Atomic} & \textbf{Composite} & \textbf{Overall} & \textbf{Rescued} \\
\midrule
\multirow{3}{*}{$\pi_{0.5}$}
& First & 41.1\% & 4.4\% & 17.6\% & \multirow{3}{*}{23} \\
& Final & 56.7\% & 10.0\% & 26.8\% & \\
& Gain & +15.6 & +5.6 & +9.2 & \\
\midrule
\multirow{3}{*}{RLDX-1}
& First & 70.0\% & 16.2\% & 35.6\% & \multirow{3}{*}{18} \\
& Final & 75.6\% & 24.4\% & 42.8\% & \\
& Gain & +5.6 & +8.1 & +7.2 & \\
\midrule
\multirow{3}{*}{WorldDreamer}
& First & 66.7\% & 15.6\% & 34.0\% & \multirow{3}{*}{21} \\
& Final & 73.3\% & 25.0\% & 42.4\% & \\
& Gain & +6.7 & +9.4 & +8.4 & \\
\bottomrule
\end{tabular}
\end{table}

Table~\ref{tab:robocasa365_agent_assisted} shows consistent gains across all three household manipulation backends. Overall success increases from 17.6\% to 26.8\% for $\pi_{0.5}$, from 35.6\% to 42.8\% for RLDX-1, and from 34.0\% to 42.4\% for WorldDreamer. These correspond to absolute gains of +9.2, +7.2, and +8.4 percentage points, respectively. The number of rescued episodes is also substantial: 23 for $\pi_{0.5}$, 18 for RLDX-1, and 21 for WorldDreamer.

The atomic/composite breakdown highlights where recovery occurs. For $\pi_{0.5}$, the largest gain appears on atomic skills, where success rises from 41.1\% to 56.7\%. For RLDX-1 and WorldDreamer, the largest gains appear on composite activities, improving by +8.1 and +9.4 points respectively. This suggests that PhyAgentOS can address both short-horizon execution errors and multi-stage household failures, depending on the failure profile of the underlying policy.

\subsubsection{Cross-Benchmark Analysis}

Across LIBERO, CALVIN, and RoboCasa365, PhyAgentOS improves final performance for every evaluated policy backend. The improvements are obtained without modifying policy weights, changing task goals, or relaxing benchmark success criteria. This supports the central premise of PhyAgentOS: many failures made by embodied policies are not fundamental task misunderstandings, but recoverable execution errors that can be identified by semantic verification and corrected through controlled continuation.

The scale and structure of the gains vary by benchmark. On LIBERO, first-attempt performance is already high for several models, so the overall gains are moderate, ranging from +0.4 to +1.3 percentage points. On CALVIN, persistent five-step task chains expose compounding errors, and PhyAgentOS produces larger full-chain improvements, including +6.7 points for $\pi_0$ and +4.1 points for $\pi_{0.5}$ on 5/5 success. On RoboCasa365, the largest absolute gains appear in household manipulation, where recovery improves overall success by +7.2 to +9.2 points across policies.

The results also show that recovery is useful at different baseline performance levels. Weaker policies, such as $\pi_0$ on CALVIN and $\pi_{0.5}$ on RoboCasa365, expose many recoverable failures and therefore benefit from larger absolute gains. Stronger policies, such as X-VLA on LIBERO and $\pi_{0.5}$ on CALVIN, leave fewer failures to recover, but PhyAgentOS still improves the hardest long-horizon metrics. This pattern indicates that the operating layer is not tied to a particular model family or success regime.

Overall, simulation-tier regression demonstrates that PhyAgentOS acts as a model-agnostic operating layer for embodied control. The policy remains responsible for action generation, while PhyAgentOS supplies semantic progress checking, failure diagnosis, and recovery from persistent physical state. The consistent gains across tabletop manipulation, chained long-horizon control, and household activity execution suggest that verifier-triggered recovery is a general mechanism for improving embodied agent reliability before real-robot deployment.

\subsection{Real-Robot Deployment and Safety Validation}

\label{sec:real_robot}

Real-robot validation tests whether the system's safety mechanisms and adapter chains function correctly under hardware noise, execution latency, and sensor uncertainty, conditions that no simulator fully captures. We deploy PhyAgentOS on a diverse fleet spanning industrial arms, desktop manipulators, dual-arm systems, and mobile platforms (Table~\ref{tab:hardware_platforms}).

{
\footnotesize
\renewcommand{\arraystretch}{1.05}
\newcommand{\cmark}{\textcolor{ForestGreen}{\checkmark}}
\newcommand{\xmark}{\textcolor{BrickRed}{$\times$}}
\captionof{table}{Supported real-robot hardware platforms. PhyAgentOS has been validated on industrial arms, desktop arms, dual-arm systems, quadruped locomotion, and humanoid robots.}\label{tab:hardware_platforms}
\vspace{-\baselineskip}
\setlength{\LTleft}{0pt plus 1fill}
\setlength{\LTright}{0pt plus 1fill}
\begin{longtable}{@{}lllccc@{}}
    \toprule
    \rowcolor{gray!25}\textbf{Manufacturer} & \textbf{Model} & \textbf{Type} & \textbf{Real Robot} & \textbf{Simulation} & \textbf{Tested} \\
    \midrule
    \endfirsthead
    \toprule
    \rowcolor{gray!25}\textbf{Manufacturer} & \textbf{Model} & \textbf{Type} & \textbf{Real Robot} & \textbf{Simulation} & \textbf{Tested} \\
    \midrule
    \endhead
    \bottomrule
    \endfoot
    \bottomrule
    \endlastfoot
    Agilex & PIPER & Robotic Arm & \cmark & \cmark & \cmark \\
    \rowcolor{gray!10}
    RealMan & RM65-B & Robotic Arm & \xmark & \cmark & \xmark \\
    \rowcolor{gray!10}
    & BOBABOT & Robotic Arm & \xmark & \cmark & \xmark \\
    \rowcolor{gray!10}
    & Elfin 5L & Robotic Arm & \xmark & \cmark & \xmark \\
    \rowcolor{gray!10}
    & Fourier GR-3 & Bipedal Humanoid & \cmark & \cmark & \xmark \\
    Franka & Franka Emika Panda & Robotic Arm & \xmark & \cmark & \xmark \\
    \rowcolor{gray!10}
    Franka & Franka FR3 & Robotic Arm & \xmark & \cmark & \xmark \\
    Unitree & G1-D & Wheeled Humanoid & \xmark & \cmark & \xmark \\
    \rowcolor{gray!10}
    Unitree & GO2 & Quadruped & \xmark & \cmark & \xmark \\
    Unitree & G1 & Bipedal Humanoid & \xmark & \cmark & \xmark \\
    \rowcolor{gray!10}
    Unitree & R1 & Bipedal Humanoid & \xmark & \cmark & \xmark \\
    Huibo & Astra-Pro & Wheeled Humanoid & \cmark & \cmark & \cmark \\
    \rowcolor{gray!10}
    Lekiwi & lekiwi & Wheeled & \cmark & \cmark & \cmark \\
    HuggingFace & SO100 & Robotic Arm & \xmark & \cmark & \xmark \\
    \rowcolor{gray!10}
    HuggingFace & SO101 & Robotic Arm & \cmark & \cmark & \cmark \\
    & Stella Gaia Hand 20 & Dexterous Hand & \cmark & \cmark & \cmark \\
    \rowcolor{gray!10}
    & ViperX300 & Robotic Arm & \xmark & \cmark & \xmark \\
    & XLerobot & Robotic Arm & \xmark & \cmark & \xmark \\
    \rowcolor{gray!10}
    & Zerith\_H1\_PRO & Wheeled Humanoid & \cmark & \cmark & \cmark \\
\end{longtable}
\par\smallskip
{\footnotesize Note: \cmark\ denotes yes and \xmark\ denotes no. Blank cells in the Manufacturer column correspond to rows where the manufacturer was not labeled in the original figure.}
}

The Franka Research 3 serves as the primary manipulation platform, used for visual reasoning, grasp execution, and ReKep~\cite{wen2024rekep} constraint-based manipulation. The AgileX PIPER and Dobot Nova 2 provide accessible desktop-arm targets for rapid iteration, with one-click deployment of ReKep and SAM3-based perception pipelines. The XLeRobot enables bimanual coordination experiments. On the mobility side, the Unitree Go2 validates locomotion adaptation, while the Unitree G1 and Zerith serve as humanoid platforms for voice-interactive task execution.

\textbf{Zero-shot cross-embodiment transfer.} A defining property of PhyAgentOS is that the same Agent protocol and \texttt{SESSIONS.md} session format drive every robot in the fleet. Changing embodiments requires swapping the \texttt{TargetAdapter} and \texttt{PolicyAdapter} in the session configuration; the cognitive layer, the \texttt{WatchdogSupervisor}, and the protocol files remain unchanged. We validate this property by executing identical task specifications across Franka, PIPER, and Dobot platforms and measuring whether the session protocol produces valid, safe trajectories on each target without embodiment-specific Agent logic.

\textbf{Safety validation.} Three safety-critical properties are evaluated on real hardware. Preflight rejection rate is measured by injecting deliberately incompatible adapter configurations and verifying that the Watchdog rejects every session before any command reaches the motor controller. SafetyGuard effectiveness is measured by injecting action chunks that violate workspace bounds or velocity limits and confirming that the guard intercepts and clamps every violation. Emergency stop latency is measured from the moment a violation is detected (at the SafetyGuard layer or the hardware emergency-stop button) to the moment the motors halt.

%% file: chapters/05_experiments/table/mc.tex
\begin{table}[h]
    \centering
    \caption{Results on the Optimus-67 Long-Horizon Benchmark~\cite{li2025optimus3}. Success Rate (SR) with standard deviation is reported across seven task groups. $\dagger$ denotes hierarchical agents where an MLLM serves as planner, and STEVE-1 acts as policy. $*$ indicates results reproduced under the same settings as other baselines.}
    \label{tab:optimus3}
    \small
    \begin{tabular}{@{}lccccccc@{}}
        \toprule
        Method & Wood & Stone & Iron & Gold & Diamond & RedStone & Armor \\
        \midrule
        \multicolumn{8}{@{}l}{\textit{Multimodal Large Language Model}$^\dagger$} \\
        \quad GPT-3.5           & 0.40$_{\pm0.15}$ & 0.20$_{\pm0.13}$ & 0.00 & 0.00 & 0.00 & 0.00 & 0.00 \\
        \quad GPT-4o            & 0.47$_{\pm0.23}$ & 0.23$_{\pm0.09}$ & 0.05$_{\pm0.04}$ & 0.00 & 0.00 & 0.00 & 0.00 \\
        \quad Gemini-1.5-pro    & 0.41$_{\pm0.14}$ & 0.21$_{\pm0.10}$ & 0.03$_{\pm0.02}$ & 0.00 & 0.00 & 0.00 & 0.00 \\
        \quad Qwen2.5-VL        & 0.28$_{\pm0.15}$ & 0.06$_{\pm0.03}$ & 0.00 & 0.00 & 0.00 & 0.00 & 0.00 \\
        \quad Qwen2.5-VL-SFT    & 0.76$_{\pm0.11}$ & 0.36$_{\pm0.07}$ & 0.11$_{\pm0.05}$ & 0.00 & 0.00 & 0.00 & 0.00 \\
        \midrule
        \multicolumn{8}{@{}l}{\textit{Goal-conditioned Policy in Minecraft}} \\
        \quad VPT               & 0.18$_{\pm0.15}$ & 0.07$_{\pm0.05}$ & 0.00 & 0.00 & 0.01$_{\pm0.01}$ & 0.00 & 0.00 \\
        \quad GROOT             & 0.34$_{\pm0.17}$ & 0.17$_{\pm0.10}$ & 0.08$_{\pm0.05}$ & 0.01$_{\pm0.01}$ & 0.01$_{\pm0.01}$ & 0.03$_{\pm0.02}$ & 0.04$_{\pm0.02}$ \\
        \quad MineCLIP          & 0.23$_{\pm0.16}$ & 0.12$_{\pm0.08}$ & 0.06$_{\pm0.05}$ & 0.00 & 0.00 & 0.00 & 0.02$_{\pm0.02}$ \\
        \quad STEVE-1           & 0.45$_{\pm0.22}$ & 0.22$_{\pm0.19}$ & 0.08$_{\pm0.06}$ & 0.00 & 0.05$_{\pm0.03}$ & 0.00 & 0.07$_{\pm0.05}$ \\
        \midrule
        \multicolumn{8}{@{}l}{\textit{Agents in Minecraft}} \\
        \quad Voyager$^*$       & 0.87$_{\pm0.25}$ & 0.32$_{\pm0.15}$ & 0.08$_{\pm0.06}$ & 0.02$_{\pm0.02}$ & 0.01$_{\pm0.01}$ & 0.00 & 0.14$_{\pm0.09}$ \\
        \quad DEPS              & 0.77$_{\pm0.13}$ & 0.48$_{\pm0.09}$ & 0.16$_{\pm0.08}$ & 0.00 & 0.01$_{\pm0.01}$ & 0.00 & 0.10$_{\pm0.18}$ \\
        \quad MP5$^*$           & 0.89$_{\pm0.23}$ & 0.73$_{\pm0.21}$ & 0.43$_{\pm0.18}$ & 0.10$_{\pm0.08}$ & 0.09$_{\pm0.08}$ & 0.17$_{\pm0.08}$ & 0.19$_{\pm0.18}$ \\
        \quad JARVIS-1          & 0.93$_{\pm0.14}$ & 0.89$_{\pm0.07}$ & 0.36$_{\pm0.06}$ & 0.07$_{\pm0.03}$ & 0.08$_{\pm0.03}$ & 0.16$_{\pm0.07}$ & 0.15$_{\pm0.19}$ \\
        \quad Optimus-1         & 0.98$_{\pm0.02}$ & 0.92$_{\pm0.04}$ & 0.46$_{\pm0.09}$ & 0.08$_{\pm0.05}$ & 0.11$_{\pm0.05}$ & 0.25$_{\pm0.03}$ & 0.19$_{\pm0.22}$ \\
        \quad Optimus-2         & 0.99$_{\pm0.02}$ & 0.93$_{\pm0.04}$ & 0.53$_{\pm0.03}$ & 0.09$_{\pm0.01}$ & 0.13$_{\pm0.02}$ & 0.28$_{\pm0.03}$ & 0.21$_{\pm0.19}$ \\
        \quad Optimus-3-Action  & 0.93$_{\pm0.03}$ & 0.87$_{\pm0.05}$ & 0.49$_{\pm0.04}$ & 0.03$_{\pm0.04}$ & 0.03$_{\pm0.02}$ & 0.09$_{\pm0.05}$ & 0.15$_{\pm0.22}$ \\
        \quad Optimus-3
            & $\mathbf{0.99}_{\pm 0.01}$
            & $0.95_{\pm 0.02}$
            & $\mathbf{0.55}_{\pm 0.03}$
            & $\mathbf{0.10}_{\pm 0.02}$
            & $0.15_{\pm 0.02}$
            & $0.29_{\pm 0.02}$
            & $\mathbf{0.23}_{\pm 0.16}$ \\

        \midrule

        \quad \textbf{PhyAgentOS$\dagger$}
            & $\mathbf{0.99}_{\pm 0.01}$
            & $\mathbf{0.96}_{\pm 0.05}$
            & $0.52_{\pm 0.19}$
            & $0.06_{\pm 0.08}$
            & $\mathbf{0.19}_{\pm 0.07}$
            & $\mathbf{0.30}_{\pm 0.16}$
            & $0.15_{\pm 0.06}$ \\
        \bottomrule
    \end{tabular}
\end{table}

%% file: chapters/05_experiments/table/star.tex
\begin{table}[htbp]
\centering
\caption{Performance comparison across different agent capabilities.}
\label{tab:star_dojo}
\setlength{\tabcolsep}{2pt}
\begin{tabular}{lcccccc}
\toprule
\textbf{Model} & \textbf{Farming} $\uparrow$ & \textbf{Crafting} $\uparrow$ & \textbf{Exploration} $\uparrow$ & \textbf{Combat} $\uparrow$ & \textbf{Social} $\uparrow$ & \textbf{Total} $\uparrow$ \\
\midrule
ReAct-like & 14.3\% & 9.5\% & 4.8\% & 0.0\% & 4.0\% & 6.7\% \\
Reflexion-like & 17.5\% & 11.9\% & 4.8\% & 2.8\% & 6.7\% & 8.7\% \\
Voyager-like & 14.3\% & 9.5\% & 6.0\% & 2.8\% & 4.0\% & 7.3\% \\
CRADLE & 25.4\% & 16.7\% & 10.7\% & 2.8\% & 8.0\% & 13.0\% \\
StarDojo & 20.6\% & 19.0\% & 9.5\% & 5.6\% & 8.0\% & 12.3\% \\
\midrule
{SPIKE (Qwen3.5-397B)} & 34.9\% & 23.8\% & 13.1\% & 8.3\% & 10.7\% & 18.0\% \\
{SPIKE (Gemini-3.1-pro)} & - & - & - & - & - & 17.7\% \\
{SPIKE (GPT5.4)} & - & - & - & - & - & 20.3\% \\
\textbf{PhyAgentOS} & 28.6\% & 50.0\% & 17.9\% & 16.7\% & 8.0\% & 22.0\% \\
\bottomrule
\end{tabular}
\end{table}

%% file: chapters/05_experiments/table/dst.tex
\begin{table}[htbp]
\centering
\caption{DeepSeek v4 Flash on \textit{Don't Starve} (Autumn Day 0, Day Phase, 10 episodes). 
PhyAgentOS enables the agent to survive twice as long and break the Day 3 barrier.}
\label{tab:dst_dojo}
\small
\setlength{\tabcolsep}{8pt}
\renewcommand{\arraystretch}{1.1}
\begin{tabular}{lccc}
\toprule
\textbf{Metric} & \textbf{Raw LLM} & \textbf{+ PhyAgentOS} & \textbf{Change} \\
\midrule
Survival Days & 1.02 $\pm$ 0.08 & \textbf{2.10 $\pm$ 0.88} & \textcolor{teal}{\textbf{+106\%}} \\
Day 3 Survival Rate & 0\% & \textbf{30\%} & \textcolor{teal}{\textbf{+30 pp}} \\
\midrule
Death by Charlie (darkness) & 90\% & 80\% & $-10$ pp \\
Death by Monster & 10\% & 10\% & --- \\
Death by Starvation & 0\% & 10\% & +10 pp \\
\midrule
Avg Health (per 10 s) & 0.870 & 0.832 & $-0.038$ \\
Avg Hunger (per 10 s) & 0.827 & 0.650 & $-0.177$ \\
Avg Sanity (per 10 s) & 0.962 & 0.880 & $-0.082$ \\
\bottomrule
\end{tabular}

\vspace{0.5em}
\footnotesize
\textbf{Note on average states:} Raw LLM dies to Charlie on Night 1 before significant resource drain; 
its high mean health/hunger/sanity reflects a \textit{short, unchallenged} lifespan. 
PhyAgentOS survives $2\times$ longer, so its lower averages naturally reflect sustained resource 
consumption over extended gameplay---evidence of prolonged survival rather than degradation.
\end{table}

%% file: chapters/06_conclusion/chapter.tex
\section{Conclusion and Future Work}
\label{sec:conclusion}

\subsection{Summary}

This paper began with a structural observation: three competing visions for AI in the physical world (VLA models, world models, and agentic systems) each capture a critical capability, yet none covers the full stack from perception through reasoning to reliable execution. Deeper fractures prevent integration. Heterogeneous hardware exposes fragmented interfaces that lock algorithms to specific embodiments. The physical execution loop offers a binary completion signal with no semantic feedback on task achievement. Architectures remain confined to single robots, single tasks, and fixed pipelines.

PhyAgentOS addresses these fractures not by building a fourth paradigm but by positioning an operating-system layer beneath the three existing ones. Its central design choice is to treat the boundary between cognition and physics as a file interface. Every cross-process datum (targets, skills, sessions, observations, lessons, and knowledge) is materialized in human-readable Markdown files with embedded YAML blocks. This \emph{State-as-a-File} protocol enables a \emph{Session-Centered Runtime} in which sessions, not atomic actions, are the unit of scheduling, preflight validation, execution, and post-hoc verification.

Four mechanisms give this architecture operational intelligence. The \texttt{SessionVerifier} replaces the binary ``action finished'' signal with semantic acceptance: an evidence bundle is evaluated against the task definition to render a verdict of success, failure, or replan, closing the open loop between intention and accomplishment. Epistemic Memory consolidates failures into \texttt{LESSONS.md} and successes into \texttt{KNOWLEDGE.md}, driving a continuous self-evolution cycle across sessions and embodiments without retraining any neural model. Benchmarking is treated as orchestration over the existing runtime surface, ensuring that every reported metric is produced by the identical execution path used in deployment. A layered safety stack (preflight contract validation, SafetyGuard at the action boundary, configurable action bridges, and watchdog-level heartbeat monitoring) provides defense in depth, with each layer remaining active even when others contain bugs.

Validation follows a progressive path that strips away confounding variables before adding them back. Game environments (Minecraft, Stardew Valley) serve as controlled-variable platforms where execution is frictionless and intelligence alone determines outcomes. Strategies validated in games are then bridged through adapter chains to simulation (adding dynamics, collision, latency) and finally to real robots (adding hardware noise, sensor uncertainty, and safety-critical constraints), with the cognitive layer unchanged throughout. The endpoint of this pipeline is always physical execution: games accelerate cognitive iteration, simulation screens for dynamics failures, but the measure of the system is whether it drives a real robot safely, reliably, and intelligently. Across three standardized benchmarks spanning this progressive ladder (Optimus-3, StarDojo, DST-Dojo), PhyAgentOS demonstrates that a unified protocol and supervisory runtime can drive diverse embodiments from game engines to industrial manipulators through the same infrastructure.

\subsection{Limitations}

Several properties of the current system bound its applicability and point toward future work.

\textbf{Polling-based protocol.} The current file-protocol implementation uses polling: the Watchdog periodically reads \texttt{SESSIONS.md} to discover new work, and the Agent polls protocol files for state updates. This design maximizes simplicity and robustness (no inter-process communication infrastructure beyond a filesystem, an append-only audit log by construction), but it introduces latency proportional to the polling interval. Event-driven watchers, such as filesystem notification hooks, could reduce this latency. However, they must preserve the single-writer semantics and atomic file updates that guarantee protocol consistency under concurrent access. A naive event-driven implementation that allows partial writes to be observed mid-update would break the protocol's correctness guarantees.

\textbf{Long-horizon memory bounds.} Epistemic Memory accumulates experience without a principled compression bound. Over thousands of sessions, \texttt{LESSONS.md} and \texttt{KNOWLEDGE.md} grow monotonically, and the \texttt{ContextBuilder} must select a relevant subset for each session's prompt. The current selection heuristic (task-type matching and recency weighting) is effective but lacks theoretical guarantees on recall of relevant past experience under adversarial task sequences. Automated skill-quality assurance also remains semi-manual: strategies are promoted to \texttt{SKILL.md} based on success-rate thresholds, but edge-case detection and regression testing of promoted skills rely on the benchmark orchestration infrastructure rather than formal verification.

\textbf{Real-robot evaluation coverage.} Real-robot experiments reported in this paper are limited to hardware-available platforms and emphasize safety-critical validation (preflight rejection, SafetyGuard effectiveness, emergency stop latency) over large-scale task-completion statistics. The full progressive pipeline (game validation to simulation regression to real-robot deployment) has been demonstrated on representative embodiments (Franka, PIPER, Dobot) but has not yet been scaled to the fleet-level diversity that PhyAgentOS's architecture theoretically supports.

\subsection{Future Work}

\textbf{Goal Graph and Session Compiler.} The current session model treats each task as an independent unit. The Goal Graph and Session Compiler, described in the architecture but not yet fully implemented, will enable DAG-level task decomposition with explicit dependency tracking across sessions, agents, and embodiments. This will allow a manipulation agent and a navigation agent to coordinate through the shared \texttt{TASK.md} protocol without custom synchronization logic, directly supporting the multi-robot coordination scenarios that real-world deployment demands.

\textbf{World model integration.} World models, treated in this paper as external reasoning resources, can be more deeply integrated into the session planning loop. A world model could provide look-ahead rollouts for the \texttt{SessionVerifier}, predicting whether a proposed session's action sequence is likely to achieve its goal before committing physical resources. On real robots, where trial-and-error carries cost and risk, this predictive capability would reduce the number of failed physical attempts and accelerate the convergence of the self-evolution loop.

\textbf{From polling to event-driven protocol.} The current polling-based protocol trades latency for simplicity. An event-driven variant, using filesystem notification hooks with guaranteed atomic-write semantics, would reduce scheduling latency while preserving the single-writer consistency that the protocol's correctness depends on. Lower latency is particularly critical for real-robot scenarios where the Watchdog must respond to safety violations within milliseconds.

\textbf{Cross-embodiment transfer at fleet scale.} The zero-shot transfer demonstrated on Franka, PIPER, and Dobot platforms validates the adapter-chain architecture at small scale. Scaling this evaluation to substantially different morphologies (legged locomotion, aerial vehicles, soft manipulators) and measuring the adapter code required per new embodiment will quantify the portability that the architecture's design implies. The long-term goal is a library of pre-validated \texttt{TargetAdapter} and \texttt{PolicyAdapter} configurations that cover the diversity of real-world robot platforms without per-embodiment retraining of the cognitive layer.

\textbf{Theoretical foundations for memory and skill quality.} Formalizing the compression bounds of Epistemic Memory (under what conditions does the \texttt{ContextBuilder} retrieve the most relevant lesson for a given task) and the correctness conditions for automated skill promotion (when does a success-rate threshold guarantee that a skill generalizes) would elevate these mechanisms from engineering heuristics to principled components. Reliable memory retrieval is a prerequisite for the self-evolution loop to function on real hardware over extended deployment periods.

\textbf{DST-Dojo: cognitive stress-testing for real-robot readiness.} The DST-Dojo benchmark, developed in concurrent work~\cite{dstdojo}, shifts evaluation from task-completion metrics to survival competence under multi-dimensional resource pressure. From the progressive validation perspective, DST-Dojo occupies the extreme end of the game tier: it forces the Agent to manage hunger, health, sanity, seasonal threats, and irreversible permadeath simultaneously. A strategy that survives 100 days in Don't Starve has demonstrated planning, risk management, and adaptation capabilities that are directly relevant to long-duration real-robot autonomy. Bridging survival-validated strategies to simulation and ultimately to physical robots through the game-to-real pipeline of Section~\ref{sec:game_agent} is a concrete path toward deploying agents that operate reliably not for minutes but for days.

%% file: chapters/07_appendix/chapter.tex
\appendix

\section{Acknowledgments}
\label{app:team}




\noindent \textbf{Contributing Members} (alphabetical order by last name)

\noindent Yicheng Chang, Ruobing Chen, Zexin Chen, Ziliang Chen, Qirui Jiao, Guanbin Li, Xiaodan Liang, Jianqi Lin, Jiongxiao Lin, Enyu Liu, Yandi Liu, Kaijun Luo, Rigenhu Mo, Shuya Mo, Yukai Shi, Yayan Tang, Yilang Tan, Hanting Wang, Yuanlei Wang, Zhijie Wang, Yifan Wen, Kaiyuan Yang, Zhanfu Yang, Kaixuan Yin, Shicheng Yin, Junhan Yu, Weipeng Zhang, Zechun Zhao, Qijun Zhong

\section{Terminology}
\label{app:terms}
\begin{itemize}
    \item Physical Agent Operating System --- PhyAgentOS
    \item Vision-Language-Action Model --- VLA
    \item Goal Planner --- Goal Planner
    \item SessionVerifier --- SessionVerifier
    \item WatchdogSupervisor --- WatchdogSupervisor
    \item SessionRunner --- SessionRunner
    \item TargetSessionHandle --- TargetSessionHandle
    \item Policy Adapter --- Policy Adapter
    \item Target Adapter --- Target Adapter
    \item Action Bridge --- Action Bridge
    \item RuntimeCompatibilityPreflight --- Preflight
    \item Epistemic Memory --- Epistemic Memory
    \item Error Taxonomy --- Error Taxonomy
    \item Directed Acyclic Graph --- DAG
\end{itemize}

\section{Protocol Schema Details}
\label{app:schema}
\begin{itemize}
    \item Full YAML schemas for the core protocol files, including \texttt{SESSIONS.md}, \texttt{SKILLRUNTIME.md}, \texttt{TARGETS.md}, \texttt{ENVIRONMENT.md}, \texttt{LESSONS.md}, \texttt{KNOWLEDGE.md}, \texttt{TASK.md}, \texttt{LOG.md}, and \texttt{RUNTIME\_SIGNAL.md}.
    \item Session state-machine transition table and error codes.
    \item \texttt{AdapterPlan} and \texttt{TargetToolManifest} structures.
    \item SessionVerifier evidence-bundle and verdict rules.
\end{itemize}

\section{Supported Target Platforms}
\label{app:profiles}
\begin{itemize}
    \item Built-in target configurations include \texttt{isaac\_remote}, \texttt{stardewvalley\_smapi}, \texttt{minecraft\_java\_env}, and \texttt{libero\_real\_remote}.
    \item Real-robot platforms: Franka Research 3, XLeRobot, AgileX PIPER, Dobot Nova 2, Unitree Go2, Unitree G1 / Zerith.
    \item Capability matrices, safety constraints, and adapter requirements per platform.
\end{itemize}